\documentclass[acmsmall, nonacm=True]{acmart}

\renewcommand\footnotetextcopyrightpermission[1]{} % removes footnote with conference information in first column
\usepackage{multirow}
\usepackage{booktabs}
\usepackage{threeparttable}

\usepackage{amsmath}

\usepackage{amssymb}
\usepackage{amsfonts}
\usepackage[caption=false,font=footnotesize]{subfig}
\usepackage{graphicx}  %插入图片的宏包
\usepackage{algorithm}
\usepackage{algorithmic}

\AtBeginDocument{%
  \providecommand\BibTeX{{%
    \normalfont B\kern-0.5em{\scshape i\kern-0.25em b}\kern-0.8em\TeX}}}

\setcopyright{acmcopyright}
\copyrightyear{2018}
\acmYear{2018}
\acmDOI{10.1145/1122445.1122456}

\begin{document}

\title{Exploring Adversarial Learning for Deep Semi-Supervised Facial Action Unit Recognition}

% \author{Shangfei Wang}
% % \authornote{Both authors contributed equally to this research.}
% \email{sfwang@ustc.edu.cn}
% \affiliation{%

\author{Shangfei Wang}
\affiliation{%
  \institution{University of Science and Technology of China}
%   \streetaddress{Hefei, Anhui 230027, PR China}
  \city{Hefei}
  \country{China}}
\email{sfwang@ustc.edu.cn}

\author{Yanan Chang}
\affiliation{%
  \institution{University of Science and Technology of China}
%   \streetaddress{Hefei, Anhui 230027, PR China}
  \city{Hefei}
  \country{China}}
\email{cyn123@mail.ustc.edu.cn}

\author{Guozhu Peng}
\affiliation{%
  \institution{University of Science and Technology of China}
%   \streetaddress{Hefei, Anhui 230027, PR China}
  \city{Hefei}
  \country{China}}
\email{gzpeng@mail.ustc.edu.cn}

\author{Bowen Pan}
\affiliation{%
  \institution{University of Science and Technology of China}
%   \streetaddress{Hefei, Anhui 230027, PR China}
  \city{Hefei}
  \country{China}}
\email{bowenpan@mail.ustc.edu.cn}

% \affiliation{%

% \affiliation{%

% \affiliation{%

% \affiliation{%



% %%
% %% By default, the full list of authors will be used in the page
% %% headers. Often, this list is too long, and will overlap
% %% other information printed in the page headers. This command allows
% %% the author to define a more concise list
% %% of authors' names for this purpose.
\renewcommand{\shortauthors}{Wang et al.}

\begin{abstract}
Current works formulate facial action unit (AU) recognition as a supervised learning problem, requiring fully AU-labeled facial images during training. It is challenging if not impossible to provide AU annotations for large numbers of facial images. Fortunately, AUs appear on all facial images, whether manually labeled or not, satisfy the underlying anatomic mechanisms and human behavioral habits. In this paper, we propose a deep semi-supervised framework for facial action unit recognition from partially AU-labeled facial images. Specifically, the proposed deep semi-supervised AU recognition approach consists of a deep recognition network and a discriminator $\mathcal{D}$. The deep recognition network $\mathcal{R}$ learns facial representations from large-scale facial images and AU classifiers from limited ground truth AU labels. The discriminator $\mathcal{D}$ is introduced to enforce statistical similarity between the AU distribution inherent in ground truth AU labels and the distribution of the predicted AU labels from labeled and unlabeled facial images. The deep recognition network aims to minimize recognition loss from the labeled facial images, to faithfully represent inherent AU distribution for both labeled and unlabeled facial images, and to confuse the discriminator. During training, the deep recognition network $\mathcal{R}$ and the discriminator $\mathcal{D}$ are optimized alternately. Thus, the inherent AU distributions caused by underlying anatomic mechanisms are leveraged to construct better feature representations and AU classifiers from partially AU-labeled data during training. Experiments on two benchmark databases demonstrate that the proposed approach successfully captures AU distributions through adversarial learning and outperforms state-of-the-art AU recognition work.
\end{abstract}




\maketitle

\section{Introduction}
Automatic facial action unit (AU) recognition has attracted increasing attention in recent years due to its wide applications in human-computer interaction. The variety of imaging conditions and differences between subjects make it challenging to correctly detect multiple facial action units from facial images. A large-scale facial image database collected from several subjects under various imaging conditions can facilitate AU classifier learning. However, current facial action unit recognition work requires fully AU-labeled facial images. AU labels must be provided by experts, and it can take hours to label a minute of video footage. It would be difficult and time-consuming, even impractical, to manually label a vast number of facial images.

\begin{figure}[!tbp]
\centering
\subfloat[co-occurrence]{\includegraphics[height=1.5in]{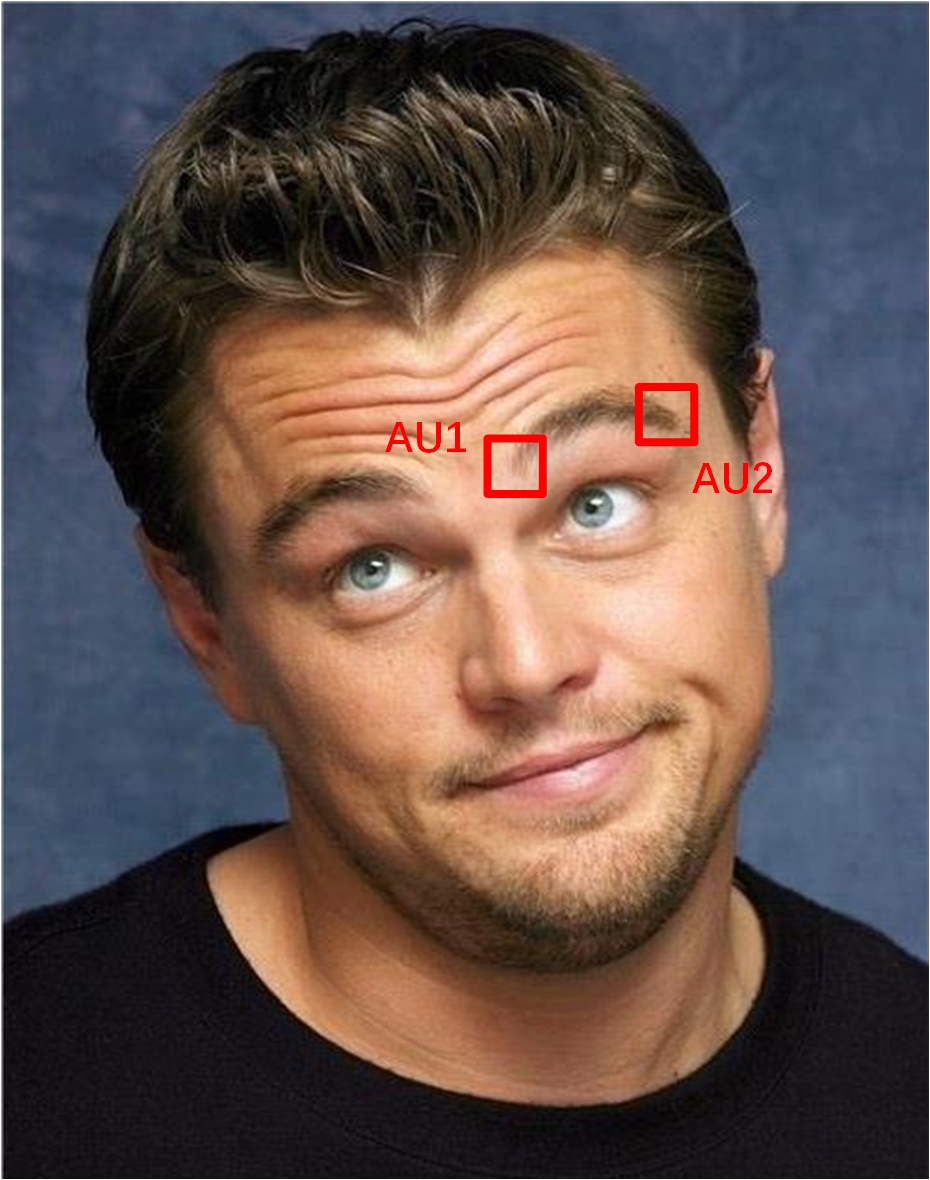}} \hspace{0.4in}
\subfloat[exclusion]{\includegraphics[height=1.5in]{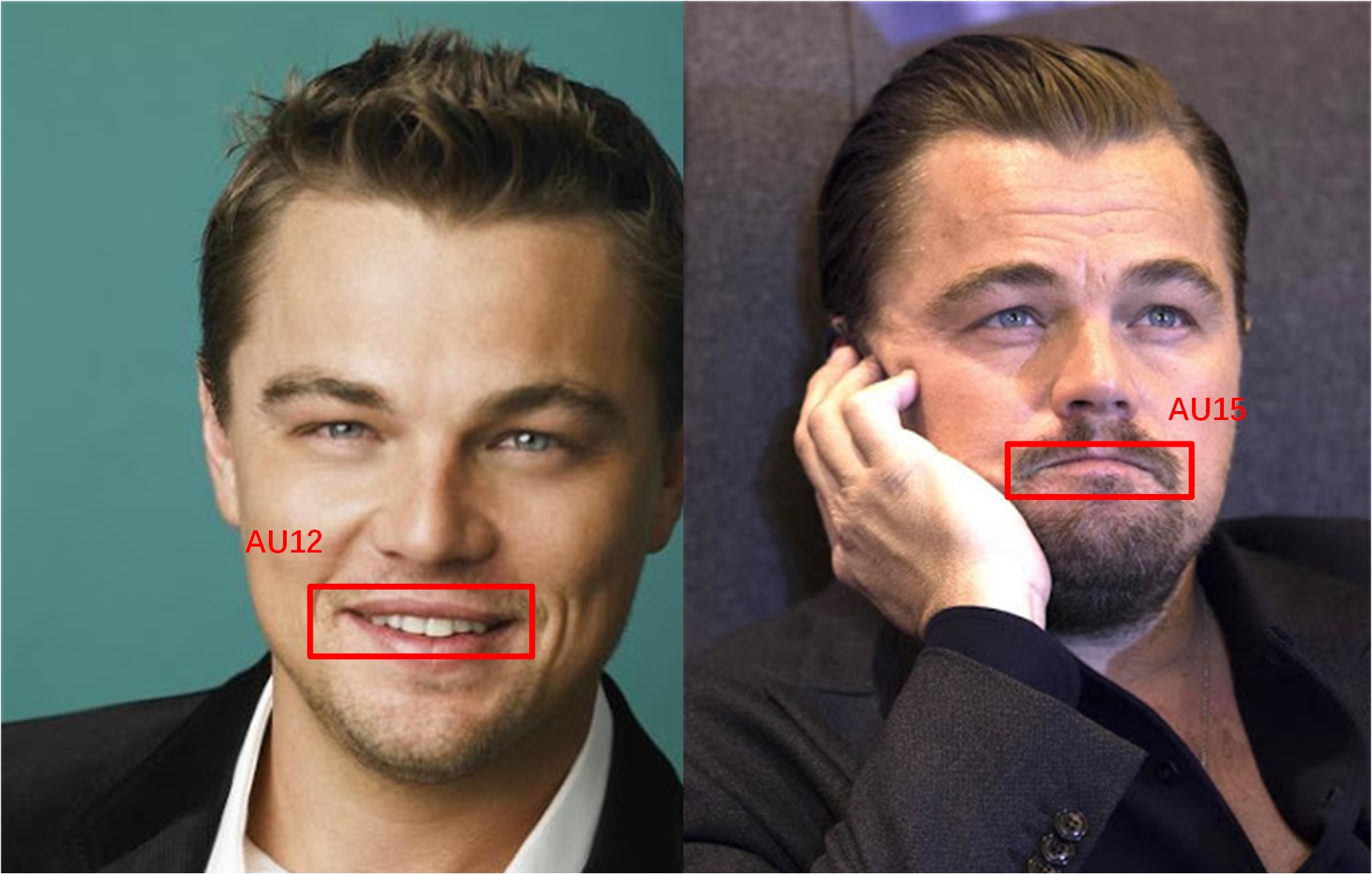}}
\caption{Examples of co-occurrence and mutually exclusive relations. (a) Both AU1 and AU2 are active. (b) Left: AU12 is active while AU15 is not active. Right: AU12 is not active while AU15 is active.}
\label{fig:examples}
\end{figure}

Facial action units describe the contraction or relaxation of one or more facial muscles. The underlying anatomic mechanisms governing facial muscle interactions lead to dependencies among AUs. For example, as shown in Figure \ref{fig:examples}, AU1 (Inner Brow Raiser) and AU2 (Outer Brow Raiser) are highly likely to appear simultaneously because AU1 and AU2 are governed by the same facial muscles, i.e., frontalis and pars medialis. Conversely, AU12 (Lip Corner Puller) and AU15 (Lip Corner Depressor) usually do not appear together, since it is anatomically impossible to represent mouth shapes of ``$\smile$'' and ``$\frown$'' at the same time. Such AU dependencies indicate that the AUs present on the face must follow certain underlying distributions. The distributions are controlled by anatomic mechanisms and thus hold true for all facial images, whether they are labeled or not.

Therefore, in this paper we propose a novel system for facial action unit recognition which learns AU classifiers from large-scale partially AU-labeled facial images, and inherent AU distributions through adversarial learning. The power of deep networks is leveraged to learn facial representations from many facial images. Then, the AU classifier $\mathcal{R}$ is trained to minimize the recognition loss from the labeled facial images and faithfully represent inherent AU distribution for both labeled and unlabeled facial images simultaneously. We further introduce a discriminator to estimate the probability that a label is the ground truth AU rather than the AU label predicted by $\mathcal{R}$. The training procedure for $\mathcal{R}$ maximizes the probability of $\mathcal{D}$ making mistakes. Through adversarial learning, we successfully exploit partial ground truth AU labels and the inherent AU distributions caused by underlying anatomic mechanisms to construct better AU classifiers from large-scale partially AU-labeled images. Experimental results on two benchmark databases demonstrate the superiority of the proposed method over state-of-the-art approaches.

\section{Related Work}
A comprehensive survey on facial action unit analysis can be found in \cite{fasel2003automatic}. This section provides a brief review of recent advances in deep AU recognition, AU recognition enhanced by label dependencies, and AU recognition from partially labeled data.

\subsection{Deep AU recognition} \label{sec:Deep AU recognition}
Due to the rapid development of deep learning in recent years, several works have utilized deep AU analysis.

Some works adopt deep networks such as convolutional neural networks (CNNs) to learn spatial representations. They then construct an end-to-end network, treating AU recognition as a multi-output binary classification problem. For example, Ghosh \emph{et al.} \cite{ghosh2015multi} and Gudi \emph{et al.} \cite{gudi2015deep} propose a multiple output CNN that learns feature representation among AUs from facial images. Khorrami \emph{et al.} \cite{khorrami2015deep} experimentally validate the hypothesis that CNNs trained for expression category recognition may be beneficial for AU analysis, since learned facial representations are highly related to AUs. Li \emph{et al.} \cite{li2018eac} introduce enhancing and cropping layers to a pre-trained CNN model. The enhancing layers are designed attention maps based on facial landmark features, and the cropping layers are used to crop facial regions around the detected landmarks and learn deeper features for each facial region. Han \emph{et al.} \cite{han2018optimizing} propose an advanced optimized filter size CNN (OFS-CNN) for AU recognition, which is capable of estimating optimal kernel size for varying image resolutions, and outperforms traditional CNNs.
Shao \emph{et al.} \cite{shao2018deep} propose a joint AU detection and face alignment framework called JAA-Net. Adaptive attention learning module is proposed to localize ROIs of AUs adaptively for better local feature extraction.
Shao \emph{et al.} \cite{shao2019facial} also propose an attention and pixel-level relation learning framework for AU recognition.
Sankaran \emph{et al.} \cite{sankaran2020domain} propose a novel feature fusion approach to combine different kinds of representations. Niu \emph{et al.} \cite{niu2019local} utilize local information and the correlations of individual local facial regions to improve AU recognition. A person-specific shape regularization method is also involved to reduce the influence of person-specific shape information.
Those works totally ignore AU dependencies while training AU classifiers.

Other works try to incorporate AU dependencies and discriminative facial regions into deep networks. For example, Zhao \emph{et al.} \cite{zhao2016deep} propose Deep Region and Multi-Label Learning (DRML) to address region learning and multi-label learning simultaneously. They adopt multi-label cross-entropy loss to model AU dependencies. As with Zhao \emph{et al.} \cite{zhao2016deep}, multi-label cross-entropy loss is adopted to model AU dependencies. Multi-label cross-entropy loss is the sum of the binary cross entropy of each label. Therefore, it does not capture label dependencies effectively.

In addition to using CNN to capture spatial representations, several works integrate long short-term memory neural networks (LSTM) to jointly capture spatial and temporal representations for AU recognition. For example, Jaiswal and Valstar \cite{jaiswal2016deep} directly combine CNNs with bidirectional long short-term memory neural networks (CNN-BiLSTM) by using the spatial representations learned from the CNN as the input of BiLSTM. Their work does not consider AU dependencies. Li \emph{et al.} \cite{li2017action} propose a deep learning framework for AU recognition that integrates region of interest (ROI) adaptation, multi-label learning, and optimal LSTM-based temporal fusing. Their multi-label learning method combines the outputs of the individual ROI cropping nets, leveraging the inter-relationships of various AUs from facial representations without considering AU dependencies among target labels. Chu \emph{et al.} \cite{chu2017learning} propose a hybrid network architecture to jointly capture spatial representations, temporal dependencies, and AU dependencies through CNN, LSTM, and multi-label cross-entropy loss respectively. As mentioned above, multi-label cross-entropy loss summarizes cross entropy of each label and does not effectively exploit label dependencies.

In summary, current deep AU recognition requires fully AU-labeled facial images. Although some methods employ multi-label cross-entropy loss to handle recognition of multiple AUs, they cannot model AU dependencies effectively since the summarization of cross entropy from each label does not represent label dependencies.

\subsection{AU recognition with learning AU relations}
Unlike the more recent interest in deep AU recognition techniques, AU recognition using shallow models has been studied for years. These shallow models explore AU dependencies through both generative and discriminative approaches.

For generative approaches, both directed graphic models (such as dynamic Bayesian networks (DBNs) \cite{tong2007facial} and latent regression Bayesian networks (LRBNs) \cite{hao2018facial}) and undirected graphic models (such as restricted Boltzmann machine (RBM) \cite{wang2013capturing}) are utilized to learn AU dependencies from ground truth AU labels. Corneanu \emph{et al.} \cite{corneanu2018deep} combine CNN and recurrent graphical model and propose Deep Structure Inference Network (DSIN) to deal with patch and multi-label learning for AU recognition. Through parameter and structure learning, graphic models can successfully capture AU distributions using their joint probabilities. 
Li \emph{et al.} \cite{li2019semantic} propose an AU semantic correlation embedded representation learning framework~(SRERL). AU knowledge-graph is utilized to guide the enhancement of facial region representation.
However, these joint probabilities have certain analytical forms and assume the inherent AU distributions follow these forms. Such an assumption may be false, since we do not have complete knowledge about AU distributions caused by muscle interactions. Furthermore, approximation algorithms are often used for the learning and inference of graphic models. This further impedes the ability of graphic models to faithfully represent AU distributions.

Discriminative approaches use additional constraints of the loss function to represent AU relations. For example, Zhu \emph{et al.} \cite{zhu2014multiple} and Zhang \emph{et al.} \cite{zhang2014simultaneous} use the constraints of multi-task learning to explore AU co-occurrences among certain AU groups. Zhao \emph{et al.} \cite{zhao2015joint} investigate the constraints of group sparsity as well as local positive and negative AU relations to select a sparse subset of facial patches and learn multiple AU classifiers simultaneously. Eleftheriadis \emph{et al.} \cite{eleftheriadis2015multi} propose constraints to encode local and global co-occurrence dependencies among AU labels to project image features onto a shared manifold.

Each of the proposed constraints can model certain AU dependencies, but none represent the hundreds of AU relations embedded in AU distributions. Furthermore, all of these works require fully AU-labeled facial images.

\subsection{AU recognition from partially labeled data}
Most current work on AU recognition formulates it as a supervised learning problem and requires fully AU-labeled facial images. Researchers have only recently begun to address AU recognition when data is partially labeled. Some of these works require expression annotations, leveraging the dependencies between AU and expressions to complement missing AU annotations. For example, Wang \emph{et al.} \cite{wang2017expression} propose a Bayesian network (BN) to capture the dependencies among AUs and expressions. For missing AU labels, the structure and parameters of the BN are learned through structured expected maximization (EM). Ruiz \emph{et al.} \cite{ruiz2015emotions} learn the AU classifier with a massive data set of expression-labeled facial images. They generate pseudo AU labels by exploiting the prior knowledge of the relations between expressions and AUs. Wang \emph{et al.} \cite{wang2018exploring} and Peng \emph{et al.} \cite{peng2018weakly} leverage an RBM and adversarial learning, respectively, to model AU relations in the task of AU recognition assisted by facial expression. Zhang \emph{et al.} \cite{zhang2018classifier} propose a knowledge-driven method to jointly learn multiple AU classifiers from images without AU annotations by leveraging prior probabilities of AUs and expressions. These works require expression labels to complement the missing AU labels.

Other methods do not require expression annotations. For example, Wu \emph{et al.} \cite{wu2015multi} and Li \emph{et al.} \cite{li2016facial} leverage the consistency between the predicted labels, the ground truth labels, and the local smoothness among the label assignments to handle the missing labels for multi-label learning with missing labels (MLML). However, the smoothness assumption is likely to be incorrect because adjacent samples in the feature space may belong to the same subject rather than the same expression. Song \emph{et al.} \cite{song2015exploiting} tackle missing labels by marginalizing over the latent values during the inference procedure for their proposed Bayesian Group-Sparse Compressed Sensing (BGCS) method. Their work handles missing labels by leveraging the characteristics of generative models, but may not outperform discriminative models in classification tasks.

Although the above works consider AU recognition under incomplete AU annotations, they all use hand-crafted features. As such, none can fully explore the advantages of deep learning when studying many facial images.

Wu \emph{et al.} \cite{wu2017deep} recently proposed a deep semi-supervised AU recognition method (DAU-R) from partially AU-labeled data and AU distributions captured by RBM. Specifically, a deep neural network is learned for feature extraction. Then, an RBM is learned to capture the inherent AU distribution using the available ground truth labels. Finally, the AU classifier is trained by maximizing the log likelihood of the AU classifiers with regard to the learned label distribution while minimizing the error between predictions and ground truth labels from partially labeled data. Their approach makes full use of both facial images with available ground truth AU labels and the massive quantity of facial images lacking annotations. However, as mentioned before, RBM is a graphic model with certain forms of joint probability. Since we do not have complete knowledge about the inherent AU distributions, the assumed analytical forms of probability may be invalid.

Therefore, in this paper we introduce an adversarial learning mechanism to learn the AU distribution directly from available ground truth labels without any assumptions of distribution form. Specifically, we introduce a discriminator $\mathcal{D}$ to distinguish the ground truth AU labels from the AU labels predicted by classifier $\mathcal{R}$. A deep network is used to learn facial representations and classifiers from both AU-labeled and unlabeled facial images by minimizing the recognition errors for the AU-labeled data and maximizing the probability of $\mathcal{D}$ making mistakes. Through an adversarial mechanism, the learned AU classifiers can minimize the predicted errors on labeled images and output AUs from any images following inherent AU distributions.

Our contribution can be summarized as follows.
(1) We are among the first to introduce adversarial learning for AU distributions, and to leverage such distributions to construct AU classifiers from partially labeled data.
(2) We conduct AU recognition experiments with completely labeled data and partially labeled data. Experimental results demonstrate the potential of adversarial learning in capturing AU distributions, and the superiority of the proposed method over state-of-the-art methods.

\section{Semi-Supervised Deep AU Recognition}
\subsection{Problem Statement}
The goal of this paper is to learn AU classifiers from a large quantity of facial images with limited AU annotations. Since certain AU distributions are controlled by anatomic mechanisms, they must be generic to all facial images regardless of annotation status. We leverage these inherent AU distributions to construct AU classifiers from partially labeled facial images. In addition to minimizing the loss between the predicted AU labels and the ground truth labels for the labeled facial images, we use adversarial learning to enforce statistical similarity between the AU distribution inherent in ground truth AU labels and the distribution of the predicted AU labels from both labeled and unlabeled facial images.

Let $T=T_{1}\cup T_{2}$ denote the training set, where $T_{1}=\{\mathbf{x}^{t}_{n},\mathbf{y}^{t}_{n}\}_{n=1}^{N}$ denotes the subset of $d$ dimensional training instances $\mathbf{x}^{t}_{n}\in\mathbb{R}^{d}$ with AU annotations and the corresponding ground truth label $\mathbf{y}^{t}_{n}\in\{1,0\}^{l}$, where $N$ is the number of instances and $l$ is the number of AU labels. $\mathbf{X}^{t}=\{\mathbf{x}^{t}_{n}\}_{n=1}^{N}$ and $\mathbf{Y}^{t}=\{\mathbf{y}^{t}_{n}\}_{n=1}^{N}$ store the labeled facial images and corresponding AU labels respectively. $T_{2}=\{\mathbf{x'}_{m}\}_{m=1}^{M}$ denotes the subset of training instances without AU annotations, where $M$ is the number of instances. $\mathbf{X}=\mathbf{X}^{t}\cup T_{2}$ stores all facial images. Given the training set with partial AU annotations, our goal is to learn an AU classifier $\mathcal{R}\colon\mathbb{R}^{d}\to\{1,0\}^{l}$ by optimizing the following formula:
\begin{equation} \label{eq:our_goal}
\min_{\Theta}(1-\alpha)\mathbb{E}_{(\mathbf{x}^{t},\mathbf{y}^{t})\thicksim T_{1}}\mathcal{L}s(\mathcal{R}(\mathbf{x}^{t}; \Theta), \mathbf{y}^{t}) + \alpha \mathbf{d}(P_{\mathbf{y},\mathbf{y'}}, P_{\mathbf{y}^{t}})
\end{equation}

where $\mathcal{L}s$ is the loss between the predicted label and the ground truth label. $\mathcal{R}$ is the AU classifier and $\Theta$ are parameters of $\mathcal{R}$. $\mathbf{y}$ and $\mathbf{y'}$ are predicted label corresponding to the labeled sample $\mathbf{x}^{t}$ and the unlabeled sample $\mathbf{x'}$ respectively. $P_{\mathbf{y},\mathbf{y'}}$ is the distribution of the predicted AU labels, and $P_{\mathbf{y}^{t}}$ is the distribution of the ground truth AU labels. $\mathbf{d}$ represents the distance between two distributions. $\alpha$ is the trade-off rate between the two terms in Equation \ref{eq:our_goal}.

In practice, it is difficult to model the distribution of the predicted AU labels ($P_{\mathbf{y},\mathbf{y'}}$), and errors may occur during the modeling procedure. To alleviate these problems, we close the two distributions through an adversarial learning mechanism.

\begin{figure*}[!tbp]
\centering
\includegraphics[width=5in]{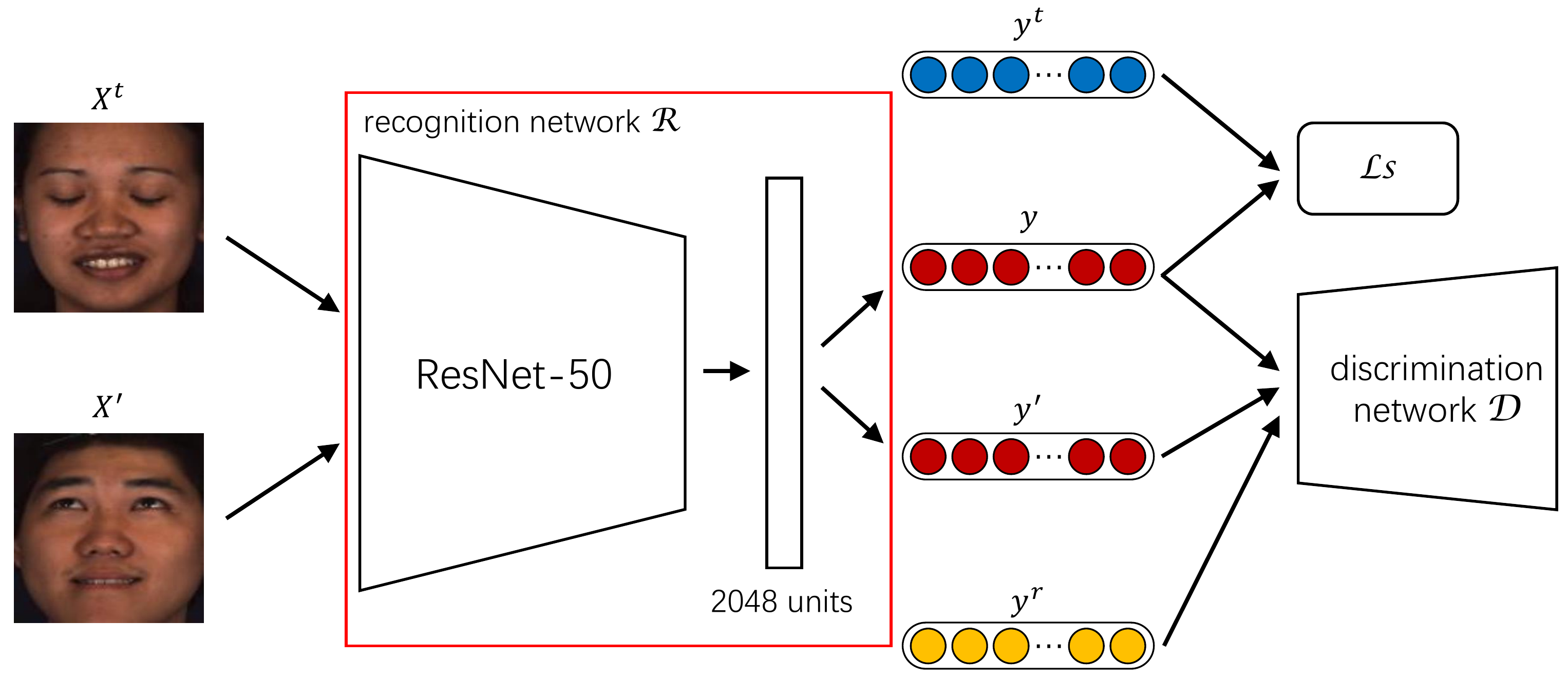}
\caption{The framework of the proposed method. A deep CNN ($\mathcal{R}$) is used as the AU classifier. Training input consists of both AU-labeled images $\mathbf{x}^t$ and unlabeled images $\mathbf{x}'$. $\mathbf{y}$ and $\mathbf{y}'$ are the predictions for the AU-labeled images and unlabeled images, respectively. The constraints for the predicted labels include the ground truth labels for labeled facial images $\mathbf{y}^t$ as well as the distribution of ground truth labels $\mathbf{y}^r$ for all inputted facial images. Specifically, $\mathbf{y}^r$ are randomly sampled from the ground truth labels set. We leverage the distribution constraints through the competition between the AU classifier and discriminator. Note that there is a one-to-one correspondence between $\mathbf{y}$ and $\mathbf{y}^t$.}
\label{fig:framework}
\end{figure*}

\subsection{Proposed Approach}

% \begin{minipage}{10cm}
\begin{algorithm}[!tp]
\caption{Adversarial semi-supervised AU recognition}
\label{alg:training}
\begin{algorithmic}[1]
\REQUIRE The partially labeled training facial images, max number of training step $K$, number of times of AU classifier updated per step $H_\mathcal{R}$, number of times of AU discriminator updated per step $H_\mathcal{D}$, sampling size $m$.
\ENSURE The AU classifier $\mathcal{R}$.
\STATE Initialize parameters of AU classifier $\Theta_{\mathcal{R}}$ and AU discriminator $\Theta_{\mathcal{D}}$.
\FOR {$k=1,2,...,K$}
\FOR {$h_{\mathcal{D}}=1,...,H_{\mathcal{D}}$}
\STATE Sample mini-batch of $m$ samples $\{\mathbf{x}_{1}, \mathbf{x}_{2},...,\mathbf{x}_{m}\} $ from facial images set $\mathbf{X}$.
\STATE Randomly sample mini-batch of $m$ labels $\{\mathbf{y}_{1}^{r}, \mathbf{y}_{2}^{r},...,\mathbf{y}_{m}^{r}\}$ from ground truth labels set $\mathbf{Y}^{t}$.
\STATE Update discriminator by descending its gradient:
\begin{equation*}
\nabla_{\Theta_{\mathcal{D}}}\Bigg[-\frac{1}{m}\sum_{i=1}^{m}\Big(\log\mathcal{D}(\mathbf{y}^{r}_{i}) + \log\left ( 1 - \mathcal{D}\left ( \mathcal{R}(\mathbf{x}_{i}) \right ) \right ) \Big)\Bigg]
\end{equation*}
\ENDFOR
\FOR {$h_{\mathcal{R}}=1,...,H_{\mathcal{R}}$}
\STATE Sample mini-batch of $m=m_{1}+m_{2}$ samples from training set $T$, which consists of $m_1$ labeled samples $\{(\mathbf{x}_{1}^{t}, \mathbf{y}_{1}^{t}), (\mathbf{x}_{2}^{t}, \mathbf{y}_{2}^{t}),...,(\mathbf{x}_{m_1}^{t}, \mathbf{y}_{m_1}^{t})\}$ from $T_1$ and $m_2$ unlabeled samples $\{\mathbf{x}^{\prime}_{1},\mathbf{x}^{\prime}_{2},...,\mathbf{x}^{\prime}_{m_2}\}$ from $T_2$.
\STATE Update the classifier by descending its gradient:
\begin{equation*}
\medmuskip=0mu
\begin{aligned}
\nabla_{\Theta_{\mathcal{R}}}&\Bigg[-\frac{\alpha}{m}\bigg(\sum_{i=1}^{m_1}\log\mathcal{D}(\mathcal{R}(\mathbf{x}_{i}^{t}))+\sum_{j=1}^{m_2}\log\mathcal{D}(\mathcal{R}(\mathbf{x}_{j}^{\prime}))\bigg)\\
&+\frac{1 - \alpha}{m_1}\sum_{i=1}^{m_1}\mathcal{L}s(\mathcal{R}(\mathbf{x}_{i}^{t}), \mathbf{y}_{i}^{t})\Bigg]
\end{aligned}
\end{equation*}
\ENDFOR
\ENDFOR
\end{algorithmic}
\end{algorithm}

We propose a novel AU recognition method leveraging adversarial learning mechanisms, inspired by Goodfellow's generative adversarial network \cite{goodfellow2014generative}. In addition to the AU classifier $\mathcal{R}$, we introduce an AU discriminator $\mathcal{D}$ to leverage the distribution constraint from limited ground truth labels. Figure \ref{fig:framework} displays the framework of the proposed method.

A training batch consisting of labeled facial images $\mathbf{x}^{t}$ and unlabeled facial images $\mathbf{x'}$ sampled from training set $T$ is inputted to AU classifier $\mathcal{R}$. The training batch is used to obtain the predicted AU labels $\mathbf{y}$ and $\mathbf{y'}$, which correspond to $\mathbf{x}^{t}$ and $\mathbf{x'}$. The predicted AU labels from $\mathcal{R}$ are regarded as ``fake'', and the ground truth $\mathbf{y}^{t}$ labels corresponding to $\mathbf{x}^{t}$ are regarded as ``real''. Both ``real'' ground truth AU labels and ``fake'' predicted AU labels are inputted to AU discriminator $\mathcal{D}$. $\mathcal{D}$ tries to distinguish the ``real'' AU labels from the predicted AU labels, while $\mathcal{R}$ tries to fool $\mathcal{D}$ into making mistakes. By leveraging the competition between $\mathcal{R}$ and $\mathcal{D}$, the distribution of the predicted AU labels nears the distribution of the ground truth AU labels until convergence. Ground truth AU labels could provide supervisory information for learning the AU classifier, minimizing the error between the predicted AU labels and the ground truth AU labels for AU-labeled samples. Therefore, we get the following objective:
\begin{equation} \label{eq:overall_objective}
\medmuskip=1mu
\begin{aligned}
\min_{\mathcal{R}}\max_{\mathcal{D}}\;&\alpha\left[\mathbb{E}_{\mathbf{y}^{r}\thicksim \mathbf{Y}^{t}}\log \mathcal{D}(\mathbf{y}^{r})+\mathbb{E}_{\mathbf{x}\thicksim \mathbf{X}}\log (1-\mathcal{D}(\mathcal{R}(\mathbf{x})))\right]\\
&+(1-\alpha)\mathbb{E}_{(\mathbf{x}^{t},\mathbf{y}^{t})\thicksim T_{1}}\mathcal{L}s(\mathcal{R}(\mathbf{x}^{t}), \mathbf{y}^{t})
\end{aligned}
\end{equation}
Observing the above equation, when $\alpha=0$, the AU classifier $\mathcal{R}$ is learned without the constraints of the label distribution.

The proposed method can be trained in a fully supervised manner if all facial images in $\mathbf{X}$ are labeled with action units. Otherwise, the proposed method is learned in a semi-supervised manner.

Similar to \cite{goodfellow2014generative}, we do not optimize Equation \ref{eq:overall_objective} directly, but use an iterative and alternative learning strategy as Equations \ref{eq:D_objective} and \ref{eq:R_objective}.
\begin{equation} \label{eq:D_objective}
\begin{aligned}
\mathcal{L}_{\mathcal{D}}=\min_{\mathcal{D}}-[&\mathbb{E}_{\mathbf{y}^{r}\thicksim \mathbf{Y}^{t}} \log\mathcal{D}(\mathbf{y}^{r})\\
& +\mathbb{E}_{\mathbf{x}\thicksim \mathbf{X}}\log (1-\mathcal{D}(\mathcal{R}(\mathbf{x})))]
\end{aligned}
\end{equation}
\begin{equation} \label{eq:R_objective}
\begin{aligned}
\mathcal{L}_{\mathcal{R}}=\min_{\mathcal{R}}-&\alpha\mathbb{E}_{\mathbf{x}\thicksim \mathbf{X}}\log \mathcal{D}(\mathcal{R}(\mathbf{x}))+\\
& (1-\alpha)\mathbb{E}_{(\mathbf{x}^{t},\mathbf{y}^{t})\thicksim T_{1}}\mathcal{L}s(\mathcal{R}(\mathbf{x}^{t}),\mathbf{y}^{t})
\end{aligned}
\end{equation}
$\mathcal{L}_{\mathcal{D}}$ is for AU discriminator $\mathcal{D}$ and $\mathcal{L}_{\mathcal{R}}$ is for AU classifier $\mathcal{R}$. In practice, it is better for $\mathcal{R}$ to minimize $-\log\mathcal{D}(\mathcal{R}(\mathbf{x}))$ instead of minimizing $\log (1-\mathcal{D}(\mathcal{R}(\mathbf{x})))$, to avoid gradient vanishing of the AU classifier \cite{goodfellow2016nips}. The detailed training procedure is described in Algorithm \ref{alg:training}.

In our work on AU recognition, we use cross-entropy loss, so $\mathcal{L}s$ can be written as Equation \ref{eq:loss_function}:
\begin{equation} \label{eq:loss_function}
\begin{aligned}
\mathcal{L}s(\mathcal{R}(\mathbf{x}^{t}),\mathbf{y}^{t})=-&\left[(\mathbf{y}^{t})^{\top}\log\mathcal{R}(\mathbf{x}^{t})+\right.\\
&\left.(\mathbf{1}-\mathbf{y}^{t})^{\top}\log(\mathbf{1}-\mathcal{R}(\mathbf{x}^{t}))\right]
\end{aligned}
\end{equation}

We use a three-layer feedforward neural network for the structure of the AU discriminator. For the AU classifier, we enable end-to-end learning directly from the input facial image via a deep CNN.
Specifically, ResNet-50 \cite{he2016deep} is used as the backbone network and one fully connected layer is built upon the 2048D feature vectors.
%The deep CNN acts only as the facial feature extractor, as shown in the red rectangle in Figure \ref{fig:framework}.

%One of the advantages of using a pre-trained model is that the network was trained on a larger database. If the network was trained in our experimental database directly, the number of training samples might be too few to learn so many parameters. Additionally, since we only update the parameters of the last three layers, it takes less time to learn the AU classifier.

% \end{minipage}

Any gradient-based method could be used to update the parameters and optimize Equations \ref{eq:D_objective} and \ref{eq:R_objective}. The Adam \cite{kingma2014adam} optimization algorithm is used in our work.

\section{Experiments}
\subsection{Experimental Conditions}
To validate the proposed adversarial AU recognition method, we conducted experiments on two benchmark databases: the BP4D database \cite{zhang2013high} and the Denver Intensity of Spontaneous Facial Action (DISFA) database \cite{mavadati2013disfa}.

The BP4D database provides both 2D and 3D spontaneous facial videos of eight facial expressions recorded from 41 subjects. Among them, 328 two-dimensional videos in which each frame is a 2D image are coded with 12 AUs: 1, 2, 4, 6, 7, 10, 12, 14, 15, 17, 23, and 24. In total, there are around 140,000 valid image samples. Like most related works, we use all available AUs and all valid samples.

\begin{figure}[!tbp]
\centering
\includegraphics[width=3.2in]{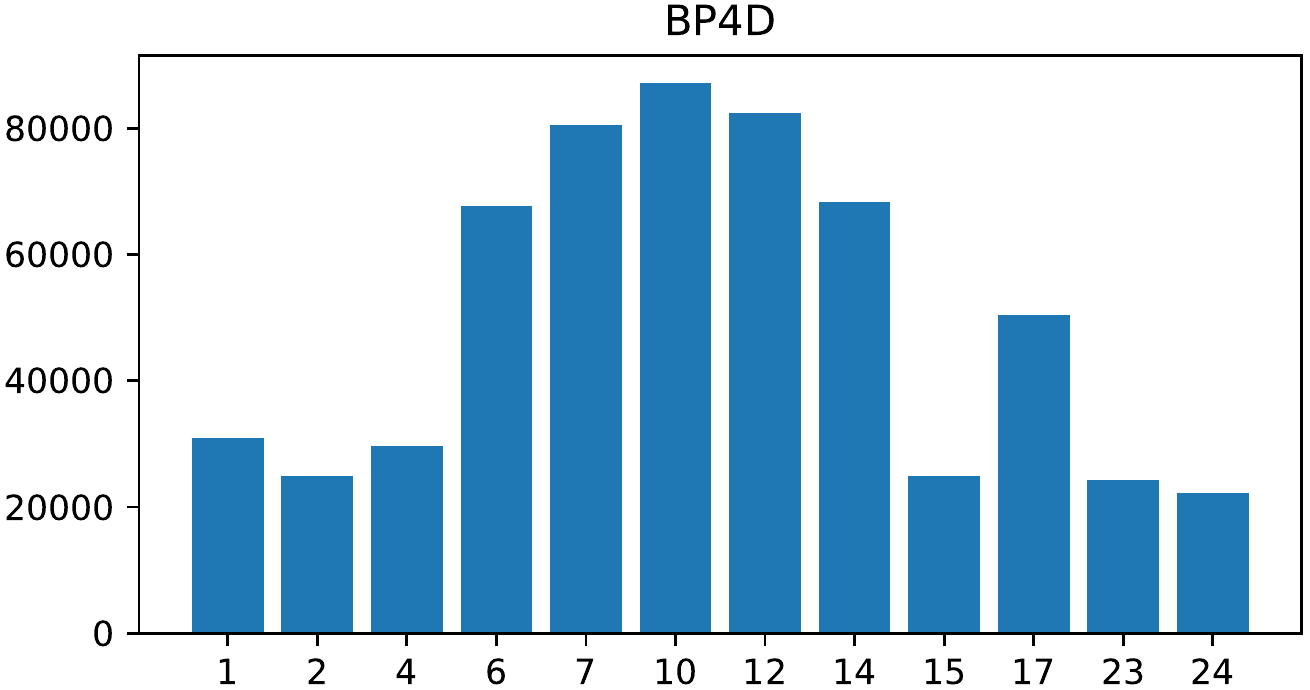}
\includegraphics[width=3.2in]{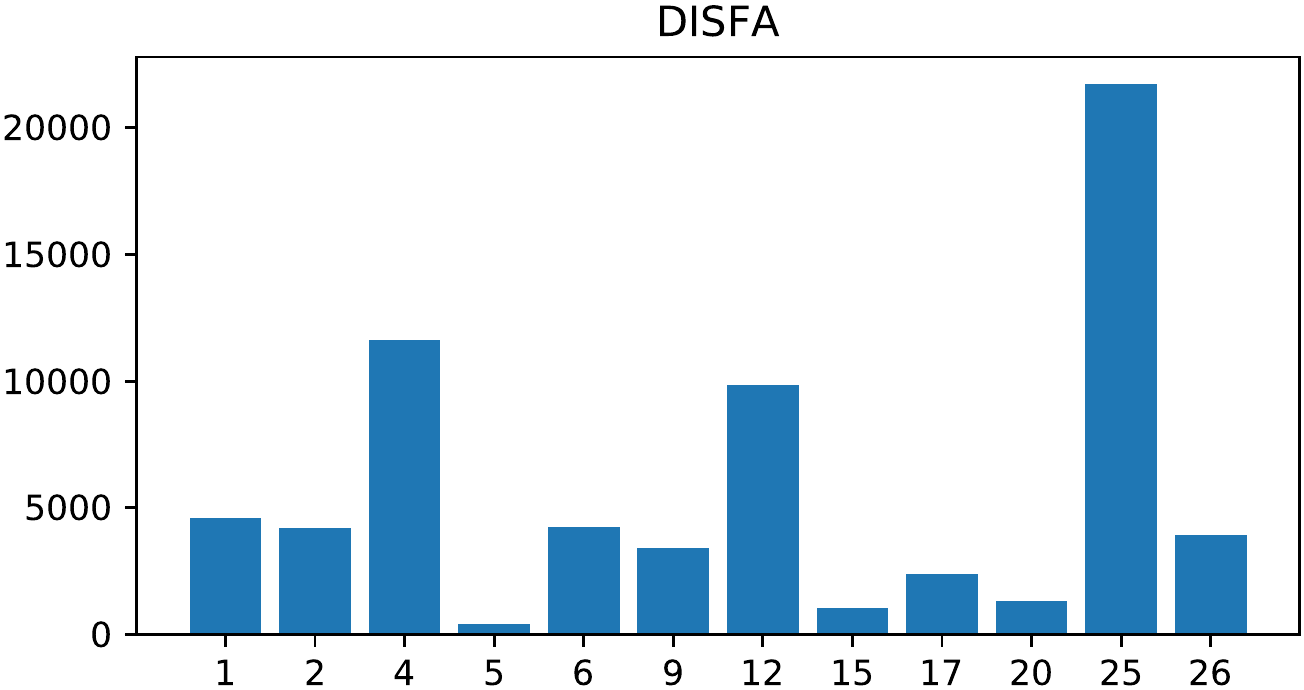}
\caption{AU occurrences of the databases.}
\label{fig:AU_occurrences}
\end{figure}

The DISFA database contains spontaneous facial videos from 27 subjects watching YouTube videos. Each image frame is coded with 12 AUs: 1, 2, 4, 5, 6, 9, 12, 15, 17, 20, 25, 26. The annotations of AUs are represented as intensities ranging from zero to five.
The number of valid image samples is around 130,000. Like most related works, we treated each AU with an intensity equal or greater than 2 as active, and considered 8 AUs (i.e., 1, 2, 4, 6, 9, 12, 25, and 26).
For both databases, the number of occurrences per AU are shown in Figure \ref{fig:AU_occurrences}.

% \begin{figure}[!htbp]
% \centering
% \subfloat{\includegraphics[width=2.9in]{resources/BP4D.pdf}} \hspace{0.1in}
% \subfloat{\includegraphics[width=2.9in]{resources/DISFA.pdf}}
% % \subfloat[exclusion]{\includegraphics[height=1.4in]{resources/example_exclusion.png}}
% \caption{AU occurrences of the databases.}
% \label{fig:AU_occurrences}
% \end{figure}

Facial alignment was taken into account to extract the face region for each image. Specifically, the face region is aligned based on three fiducial points: the centers of each eye and the mouth. After cropping and warping the face region, all images in both databases were normalized to $224\times224\times3$ pixels.

We conduct semi-supervised AU recognition experiments on images with incomplete AU annotations and fully supervised AU recognition experiments on completely AU-annotated images on both databases. For semi-supervised scenarios, we randomly miss AU labels according to certain proportions: 10\%, 20\%, 30\%, 40\%, and 50\%. On the BP4D database, we adopt three-fold subject-independent cross validation, a commonly used experimental strategy for AU recognition.
For a fair comparison to \cite{wu2017deep} and \cite{han2018optimizing}, we apply their AU selection and data split strategy on the BP4D database. On the DISFA database, similar to most related works, we adopt 3-fold subject-independent cross validation. Because of the different experimental conditions with \cite{wu2017deep} and \cite{han2018optimizing}, we don't conduct 9-Folds and 10-Folds experiments on the DISFA database. We adopt F1 score, AUC and accuracy to evaluate the performance of the proposed method. 
Moreover, the best values of alpha in the above-mentioned experiments are the same of the optimal values in Section 4.3.2. %%需要修改

To demonstrate the effectiveness of the adversarial mechanism in the semi-supervised scenario, we compare our method to the method without label constraint (``O-wlc''
for short). For this method, we set the tradeoff rate $\alpha=0$ in Equation \ref{eq:overall_objective} to remove the AU label distribution constraint.
For the semi-supervised scenario, we also compare the proposed method to DAU-R \cite{wu2017deep} and BGCS \cite{song2015exploiting}. We re-conduct the semi-supervised experiments of BGCS using their provided codes, since Song \emph{et al.} \cite{song2015exploiting} didn't conduct semi-supervised experiments on the BP4D or DISFA databases. Totally, for experiments on the DISFA database, we compare our method to BGCS and O-wlc. DAU-R is not involved due to the difference of experimental conditions.  And O-wlc, DAU-R and BGCS are employed to  compare with our method on the BP4D database. In addition, we don't compare our results with MLML\cite{wu2015multi} on both databases because of the memory problems. We do not compare our method to the methods found in \cite{ruiz2015emotions,wang2017expression,wang2018exploring,peng2018weakly} and \cite{zhang2018classifier} either, since these works require expression annotations.

% To demonstrate the effectiveness of the adversarial mechanism in the fully supervised scenario, we compare our method to the method without label constraint (``O-wlc''
% for short). For this method, we set the tradeoff rate $\alpha=0$ in Equation \ref{eq:overall_objective} to remove the AU label distribution constraint.
For the fully supervised scenario, we compare the proposed method to state-of-the-art works, including ARL \cite{shao2019facial}, LP-Net \cite{niu2019local}, SRERL \cite{li2019semantic}, U-Net \cite{sankaran2020domain}, DSIN \cite{corneanu2018deep}, JAA-Net \cite{shao2018deep}, EAC-Net \cite{li2018eac}, OFS-CNN \cite{han2018optimizing}, DAU-R \cite{wu2017deep}, CPM \cite{zeng2015confidence}, DRML \cite{zhao2016deep}, JPML\cite{zhao2015joint}, and APL\cite{zhong2014learning}.
The results of CPM are taken from \cite{chu2017learning}, since Zeng \emph{et al.} \cite{zeng2015confidence} did not provide the results on three-fold protocol.
The experiments conducted in Hao \emph{et al.}'s work \cite{hao2018facial} are based on the apex facial images with expression labels, and \cite{li2017action} and \cite{chu2017learning} use temporal models, so these methods are ignored in the comparison.
In \cite{li2017action}, the best results from non-temporal model (ROI) are used for comparison.

%In addition to within-database experiments, we conduct cross-database experiments to evaluate the generalization ability of our method. Due to the huge difference in the number of samples between the two databases, we train on the BP4D database and test on the DISFA database. Both fully supervised and semi-supervised cross-database experiments are conducted. We compare the proposed method to ``O-wlc'' for both fully supervised and semi-supervised cross-database experiments. In the fully supervised scenario, we also compare our method to DRML \cite{zhao2016deep}, the only related work that also conducted cross-database experiments by training on the BP4D database and testing on the DISFA database. In the semi-supervised scenario, we randomly miss AU labels on the training database according to certain proportions (i.e., 10\%, 20\%, 30\%, 40\%, and 50\%). The experiment was repeated five times for each proportion to alleviate the effects of randomness. Average F1 score is reported as the experiment metric.

\subsection{Semi-Supervised AU recognition}
\subsubsection{Experimental Results of Semi-Supervised AU Recognition}

\begin{figure}[!tbp]
\centering
\includegraphics[width=3.3in]{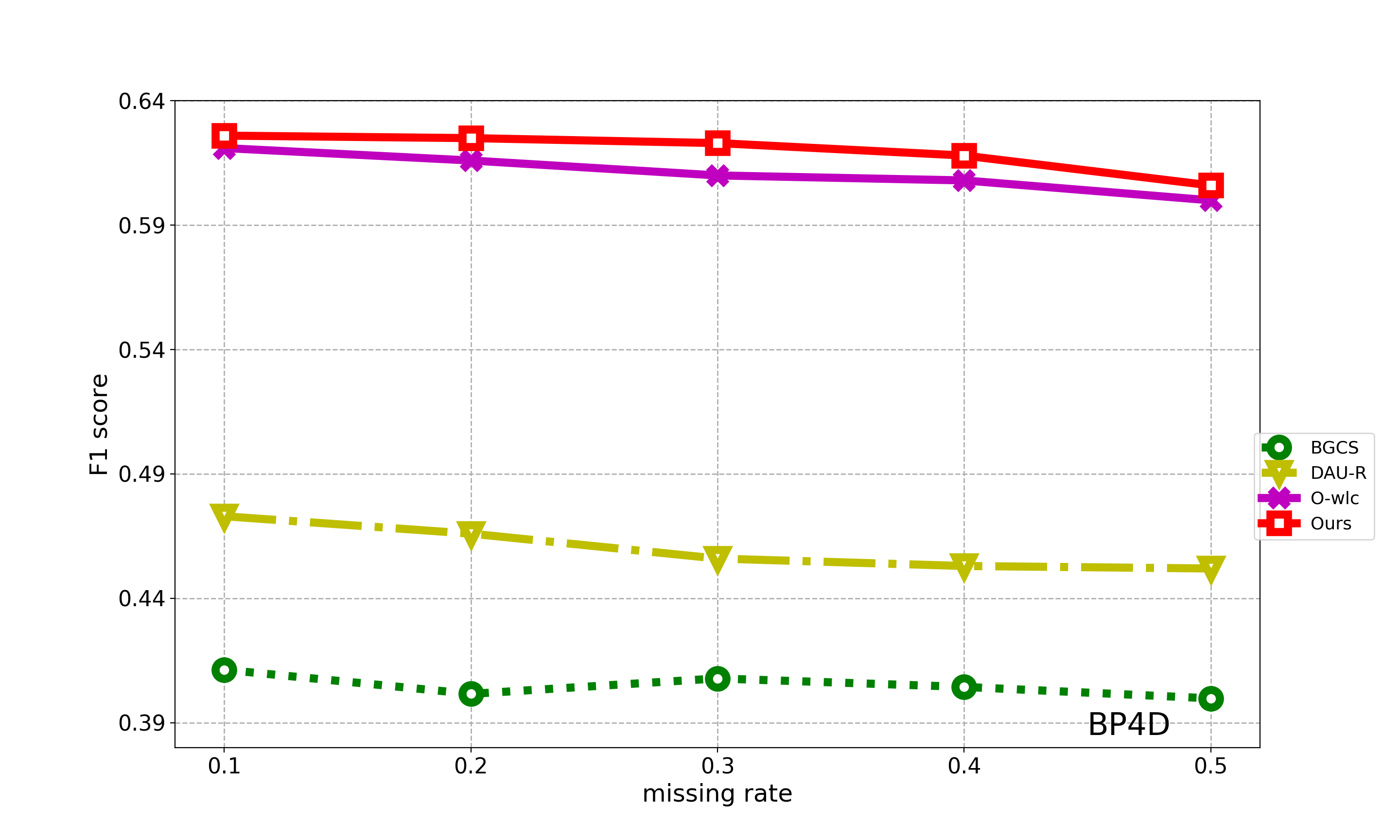}
\includegraphics[width=3.3in]{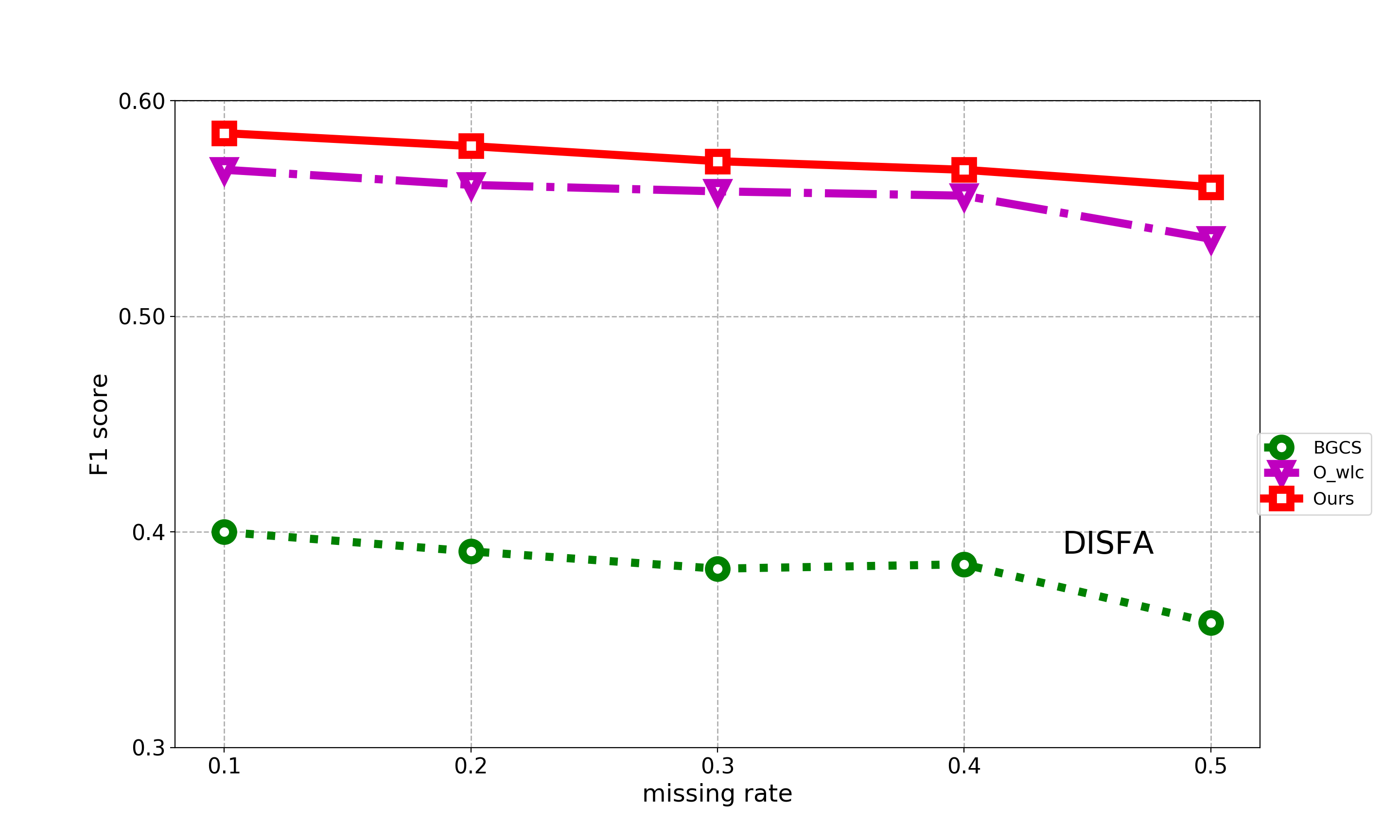}
\caption{Experimental results of semi-supervised AU recognition. Up: results on the BP4D database. Down: results on the DISFA database.}
\label{fig:semi_results}
\end{figure}

Experimental results of semi-supervised AU recognition are shown in Figure \ref{fig:semi_results}. 
% From Figure \ref{fig:semi_results}, 
We can observe the following:

First, the performances of most methods illustrated in Figure \ref{fig:semi_results} maintain a downward trend as the missing rate increases. For example, on the BP4D database, the average F1 scores of our method drop from 62.6\% in 0.1 missing rate to 60.6\% in 0.5 missing rate. On the DISFA database, the average F1 scores of our method drop from 58.5\% to 56.0\% as the missing rate increases. This is expected, since additional ground truth AU labels provide more supervisory information to better train AU classifiers.

Secondly, when using different missing rates, the proposed method substantially outperforms the ``O-wlc'' method, with higher average F1 scores on both databases. The adversarial learning mechanism introduced in our method leverages dependencies among AUs so that unlabeled facial images also improve the classifier. This leads to better results in the semi-supervised learning scenario.

Thirdly, compared to DAU-R, which is also a deep AU recognition approach using partially labeled images, the proposed method achieves a higher F1 score for all missing rates on the BP4D database. Instead of using an RBM, the proposed method successfully leverages an adversarial mechanism to capture the statistical distribution of AU labels. This ensures that AUs appearing on all images follow statistical distributions based on facial anatomy and humans' behavioral habits. Therefore, the proposed method thoroughly takes advantage of limited ground truth AU labels and readily available large-scale facial images to achieve better performance than DAU-R in the semi-supervised learning scenario.

Fourthly, the proposed method outperforms the shallow weakly supervised method(BGCS), with higher F1 scores for all missing rates on both databases respectively. This is an additional evidence of the strength of deep learning and the competence of adversarial mechanisms in semi-supervised learning scenarios.

\subsubsection{Evaluation of Adversarial Learning}

\begin{figure}[!htbp]
\centering
\subfloat[The marginal distribution of each AU in predicted AU labels with different methods and the marginal distribution in ground truth AU labels.]{\includegraphics[width=2.8in]{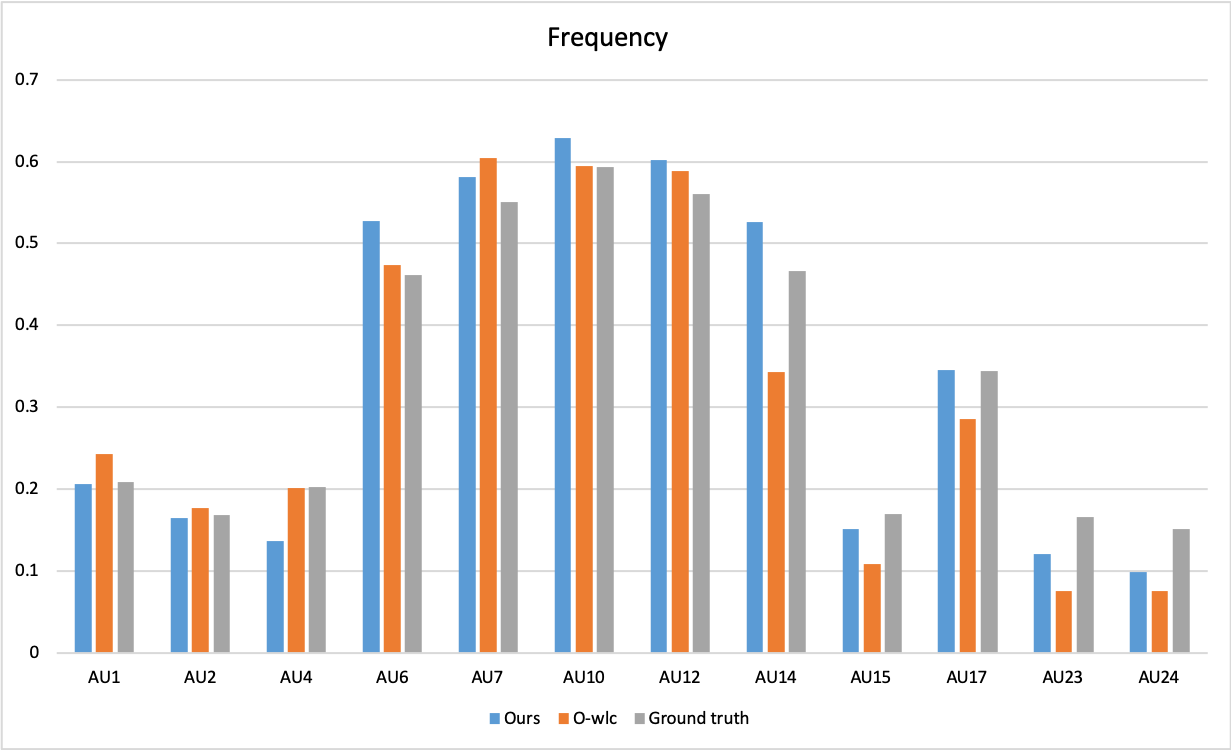}\label{fig:frequency}}
% \hfill
\hspace{0.3in}
\subfloat[The absolute difference between the marginal distributions of each predicted AU label and the ground truth AU labels.]{\includegraphics[width=2.8in]{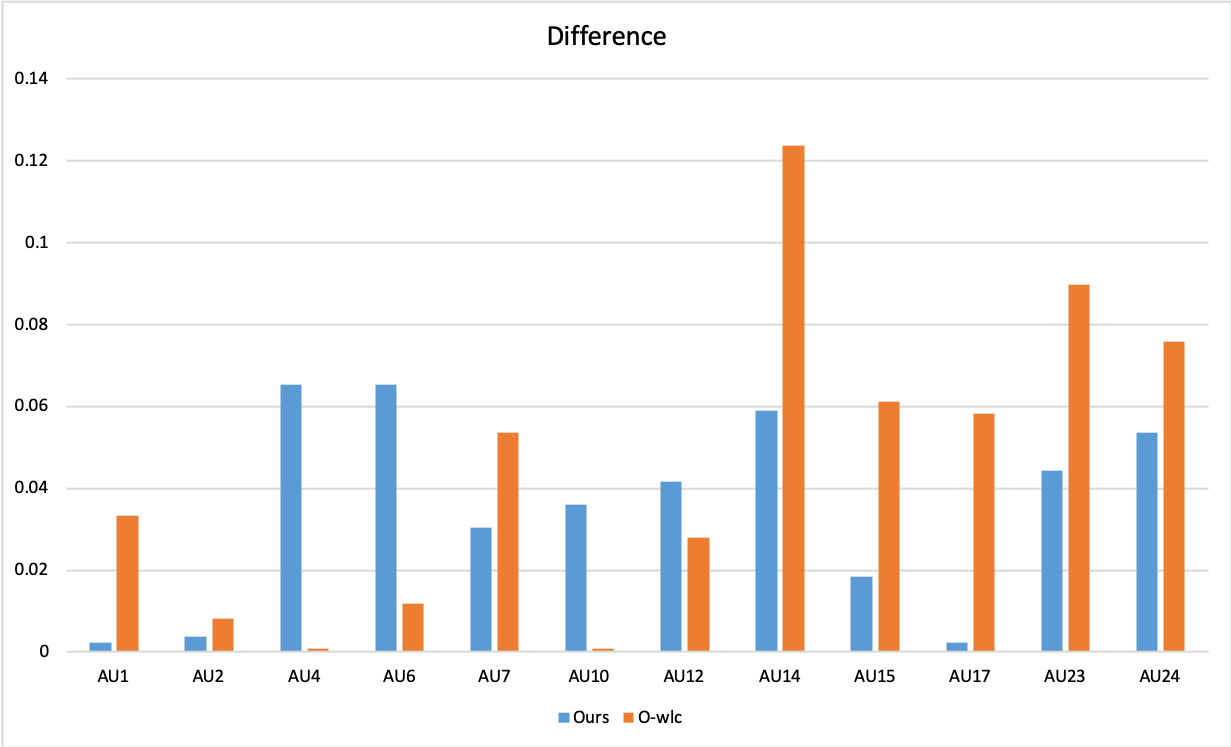}\label{fig:difference}}
\caption{Evaluation of adversarial learning on the BP4D database in terms of marginal distributions.}
\end{figure}

\begin{figure}[!thbp]
\centering
% \subfloat[DAU-R]{\includegraphics[width=3.4in]{resourcDAU-R.png}}
\subfloat[``O-wlc'']{\includegraphics[width=3.3in]{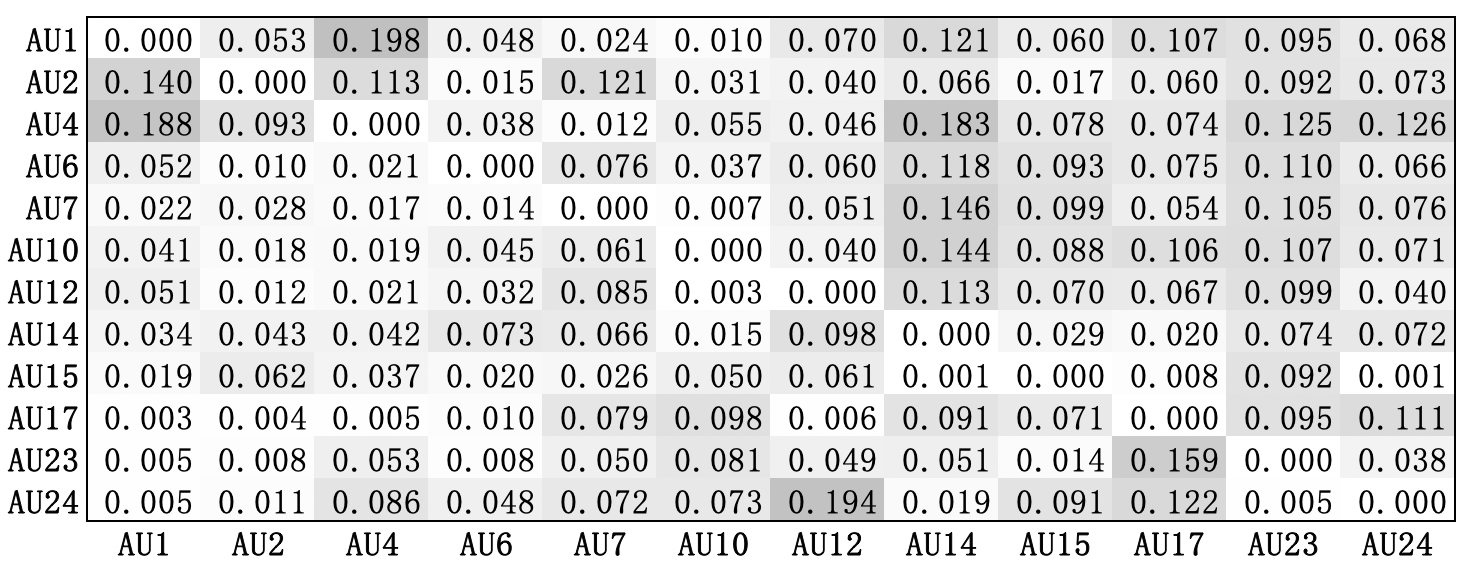}}
\hspace{0.4in}
\subfloat[``Ours'']{\includegraphics[width=3.3in]{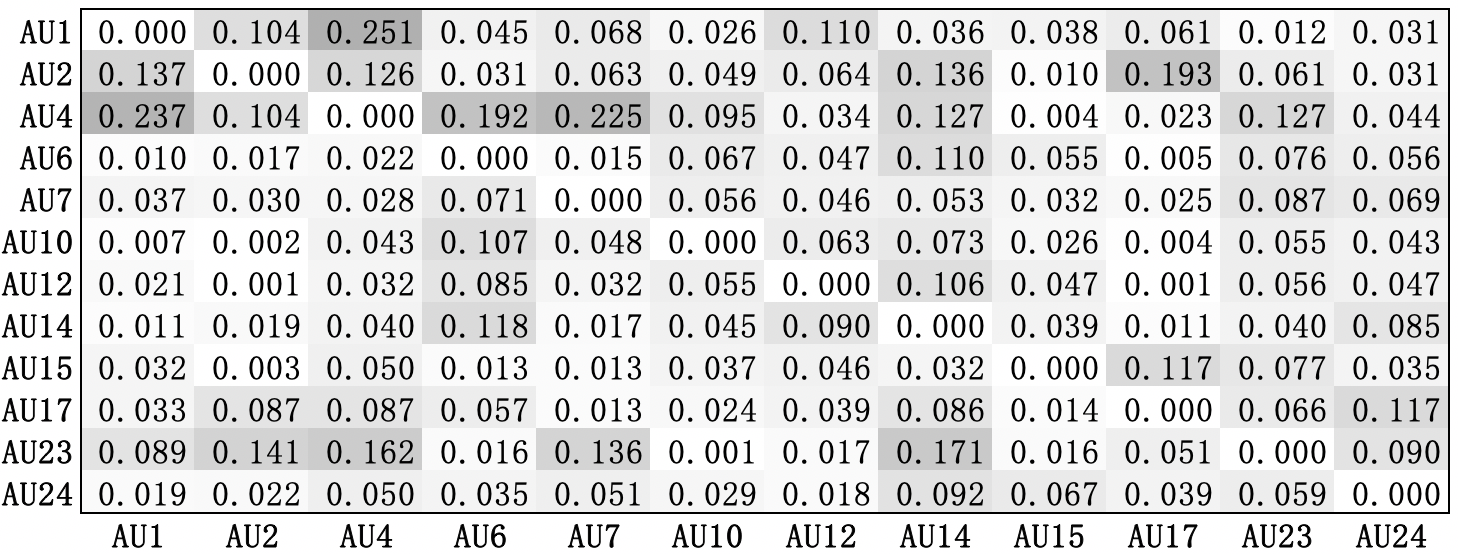}}
% \subfloat[Ours]{\includegraphics[width=3.4in]{resources/ours.png}}
\caption{Evaluation of adversarial learning on the BP4D database in terms of conditional distributions.}
\label{fig:condition}
\end{figure}

\

To validate the effectiveness of adversarial learning for capturing statistical distributions of multiple AUs from ground truth AU labels, we calculate the marginal and conditional distributions of each AU in the predicted AU labels of the testing set, and compare them to the distribution of each AU in the ground truth labels. 
For example, Figure \ref{fig:frequency} illuminates the marginal distributions of the predicted AUs from the proposed method and O-wlc, as well as the distributions of each AU in the ground truth labels for the BP4D database. 
It should be noted that we discuss the results of ours and O-wlc in 0.5 missing rate here.
% \textbf{In particular, we also compare the marginal distributions of the predicted AUs in the semi-supervised scenario. The results in 0.5 missing rate are involved("Semi" for short).} 
Figure \ref{fig:difference} shows the absolute differences between the marginal distribution of each AU in the predicted AU labels and the marginal distribution of each AU in the ground truth AU labels for these two methods. Figure \ref{fig:condition} shows the absolute differences between the conditional distribution of one AU under another AU in the predicted AU labels, and the conditional distributions in ground truth AU labels.

From Figure \ref{fig:difference} and Figure \ref{fig:condition}, we find that the marginal and conditional AU distributions predicted by ours are closest to the AU distribution of the ground truth AU labels. As seen in Figure \ref{fig:difference}, the absolute differences between ours and ground truth are lower for eight AUs than ``O-wlc''. Figure \ref{fig:condition} shows that the average absolute differences of conditional distributions between ground truth and predicted AU are 0.054 and 0.056 for ours and ``O-wlc'', respectively. The proposed method outperforms ``O-wlc'' with lower absolute differences of conditional and marginal distributions. ``O-wlc'' minimizes the recognition errors of each AU but ignores AU dependencies, while the proposed method leverages adversarial learning to capture AU distributions. Therefore, the proposed method can successfully make the distribution of predicted AU labels closer to the distribution of ground truth AU labels. It demonstrates the effectiveness of the adversarial learning for leveraging statistical distributions of multiple AUs from ground truth AU labels.

\subsection{Fully Supervised AU Recognition}
\subsubsection{Experimental Results of Fully Supervised AU Recognition}

\

The fully supervised experimental results of AU recognition on the BP4D and DISFA database are listed in Tables \ref{table:bp4d_f1}, \ref{table:bp4d_auc_acc} and  \ref{table:disfa_f1_auc_acc}.

%%% �Ա�O-wlc
The proposed method performs better than the baseline method ``O-wlc'' on both databases. Specifically, on the BP4D database, the proposed method achieves 1.3\% and 0.9\% higher average F1 score than ``O-wlc'' on 60\%-20\%-20\% and three-fold protocol, and a 0.7\% higher average AUC score than ``O-wlc'' on three-fold protocol. On the DISFA database, our method achieves a 1.7\% higher average F1 score than ``O-wlc''.  
Unlike ``O-wlc'', which only minimizes the error of the predicted AUs and ground truth AUs, the proposed method imposes adversarial loss to enforce statistical similarity between the predicted and ground truth AUs. Thus, the proposed method can leverage AU distributions existing in ground truth AU labels to construct better AU classifiers, obtaining better performance on AU recognition.

To further demonstrate the superiority of the proposed method, we compare our method to several state-of-the-art works. From Table \ref{table:bp4d_f1}-\ref{table:disfa_f1_auc_acc}, we observe the following:

%%%%%%%%%%%%%%%%%%%%%%%%%%%%%%%%% BP4D F1 score %%%%%%%%%%%%%%%%%%%%%%%%%%%%%%%%%
\begin{table}[!htbp]
\caption{F1 score of fully supervised AU recognition experiments on the BP4D database.}
\label{table:bp4d_f1}
\centering
\begin{threeparttable}
\tabcolsep=3pt
\scalebox{0.7}[0.75]{
\begin{tabular}{c|ccc|ccc|ccccccccccccc}\toprule
& \multicolumn{3}{c|}{\textbf{training/development F1}} & \multicolumn{3}{c|}{\textbf{60\%-20\%-20\% F1}} & \multicolumn{9}{c}{\textbf{3-fold F1}} \\ \cmidrule{2-20}
AU & Ours & ``O-wlc'' & \begin{tabular}[c]{@{}c@{}}\scriptsize OFS-CNN\\\cite{han2018optimizing}\end{tabular} &
Ours & ``O-wlc'' & \begin{tabular}[c]{@{}c@{}}\scriptsize DAU-R\\\cite{wu2017deep}\end{tabular} &
Ours & ``O-wlc'' & 
\begin{tabular}[c]{@{}c@{}}\scriptsize SRERL\\\cite{li2019semantic}\end{tabular} &
\begin{tabular}[c]{@{}c@{}}\scriptsize ARL\\\cite{shao2019facial}\end{tabular} &
\begin{tabular}[c]{@{}c@{}}\scriptsize LP-Net\\\cite{niu2019local}\end{tabular} &
\begin{tabular}[c]{@{}c@{}}\scriptsize U-Net\\\cite{sankaran2020domain}\end{tabular} &
\begin{tabular}[c]{@{}c@{}}\scriptsize DSIN\\\cite{corneanu2018deep}\end{tabular} &
\begin{tabular}[c]{@{}c@{}}\scriptsize JAA-Net\\\cite{shao2018deep}\end{tabular} &
\begin{tabular}[c]{@{}c@{}}\scriptsize ROI\\\cite{li2017action}\end{tabular} &
\begin{tabular}[c]{@{}c@{}}\scriptsize EAC-Net\\\cite{li2018eac}\end{tabular} &
\begin{tabular}[c]{@{}c@{}}\scriptsize CPM\\\cite{zeng2015confidence}\end{tabular} &
\begin{tabular}[c]{@{}c@{}}\scriptsize DRML\\\cite{zhao2016deep}\end{tabular} &
\begin{tabular}[c]{@{}c@{}}\scriptsize JPML\\\cite{zhao2015joint}\end{tabular} \\ \midrule
1    & \textbf{41.9} & 32.4 & 41.6 & 48.7 & \textbf{51.6} & 38.1 & 50.8 & 48.6 & 46.9 & 45.8 & 43.4 & 49.1 & \textbf{51.7} & 47.2 & 36.2 & 39.0 & 43.4 & 36.4 & 32.6 \\
2    & \textbf{37.9} & 32.1 & 30.5 & \textbf{28.0} & 23.0 & 16.6 & 44.7 & 42.0 & \textbf{45.3} & 39.8 & 38.0 & 44.1 & 41.6 & 44.0 & 31.6 & 35.2 & 40.7 & 41.8 & 25.6 \\
4    & \textbf{50.9} & 46.2 & 39.1 & \textbf{56.0} & 55.5 & 41.8 & 56.8 & \textbf{59.4} & 55.6 & 55.1 & 54.2 & 50.3 & 58.1 & 54.9 & 43.4 & 48.6 & 43.3 & 43.0 & 37.4 \\
6    & \textbf{78.6} & 78.5 & 74.5 & \textbf{76.6} & 76.3 & 74.1 & \textbf{79.6} & 78.4 & 77.1 & 75.7 & 77.1 & 79.2 & 76.6 & 77.5 & 77.1 & 76.1 & 59.2 & 55.0 & 42.3 \\
7    & 77.6 & \textbf{79.4} & 62.8 & \textbf{76.4} & 76.0 & 62.0 & 80.6 & \textbf{80.8} & 78.4 & 77.2 & 76.7 & 74.7 & 74.1 & 74.6 & 73.7 & 72.9 & 61.3 & 67.0 & 50.5 \\
10   & \textbf{84.3} & 83.8 & 74.3 & 88.0 & \textbf{88.6} & 73.9 & 84.1 & 84.5 & 83.5 & 82.3 & 83.8 & 80.9 & \textbf{85.5} & 84.0 & 85.0 & 81.9 & 62.1 & 66.3 & 72.2 \\
12   & \textbf{86.6} & 85.4 & 81.2 & \textbf{83.0} & 82.6 & 79.2 & \textbf{88.4} & 88.3 & 87.6 & 86.6 & 87.2 & 88.3 & 87.4 & 86.9 & 87.0 & 86.2 & 68.5 & 65.8 & 74.1 \\
14   & \textbf{65.5} & 62.7 & 55.5 & 56.1 & 58.2 & \textbf{58.8} & 66.8 & 63.1 & 63.9 & 58.8 & 63.3 & 63.9 & \textbf{72.6} & 61.9 & 62.6 & 58.8 & 52.5 & 54.1 & 65.7 \\
15   & 43.5 & \textbf{44.4} & 32.6 & \textbf{41.4} & 34.8 & 24.6 & 52.0 & 51.1 & \textbf{52.2} & 47.6 & 45.3 & 44.4 & 40.4 & 43.6 & 45.7 & 37.5 & 36.7 & 33.2 & 38.1 \\
17   & 60.3 & \textbf{60.7} & 56.8 & 61.4 & \textbf{64.2} & 56.4 & 60.4 & 59.7 & 63.9 & 62.1 & 60.5 & 60.3 & \textbf{66.5} & 60.3 & 58.0 & 59.1 & 54.3 & 48.0 & 40.0 \\
23   & \textbf{44.0} & 34.0 & 41.3 & \textbf{49.7} & 38.4 & 26.1 & 47.7 & 46.4 & 47.1 & 47.4 & \textbf{48.1} & 41.4 & 38.6 & 42.7 & 38.3 & 35.9 & 39.5 & 31.7 & 30.4 \\
24   &  -   &  -   &  -   & 38.7 & \textbf{39.0} & 37.6 & 47.6 & 46.4 & 53.3 & \textbf{55.4} & 54.2 & 51.2 & 46.9 & 41.9 & 37.4 & 35.8 & 37.8 & 30.0 & 42.3 \\ \midrule
Avg. & \textbf{61.0} & 58.1 & 53.7 & \textbf{58.7} & 57.4 & 49.1 & \textbf{63.3} & 62.4 & 62.9 & 61.1 & 61.0 & 60.6 & 61.7 & 60.0 & 56.4 & 55.9 & 50.0 & 48.3 & 45.9 \\
\bottomrule
\end{tabular}}
\end{threeparttable}
\end{table}

% 'alpha': 0.01  Final
% mean f1=0.632951, auc=0.728992, acc=0.786243
% f1: [0.50782, 0.44712, 0.56772, 0.7963, 0.8057, 0.84133, 0.88366, 0.66801, 0.52009, 0.60416, 0.47712, 0.47638]
% auc: [0.7036, 0.67834, 0.74361, 0.80193, 0.75016, 0.76768, 0.8626, 0.66756, 0.70833, 0.69267, 0.68981, 0.68161]
% acc: [0.74112, 0.7831, 0.80095, 0.79788, 0.76411, 0.7952, 0.86712, 0.66378, 0.84087, 0.7031, 0.8213, 0.85638]

% 'alpha' : 0 Final mean f1=0.623866, auc=0.722355, acc=0.784419 
% f1: [0.48609, 0.41984, 0.59419, 0.78405, 0.80789, 0.84477, 0.88281, 0.63106, 0.51091, 0.59722, 0.46391, 0.46365] 
% auc: [0.68439, 0.65732, 0.75337, 0.78501, 0.74765, 0.78667, 0.86025, 0.64941, 0.70167, 0.68637, 0.68363, 0.67252]
% acc: [0.7393, 0.77997, 0.82571, 0.77851, 0.7634, 0.80595, 0.86549, 0.64995, 0.83986, 0.6957, 0.81234, 0.85685]

%%%%%%%%%%%%%%%%%%%%%%%%%%%%%%%%% BP4D AUC Accuracy %%%%%%%%%%%%%%%%%%%%%%%%%%%%%%%%%
\begin{table*}[!htbp]
\caption{AUC and accuracy of fully supervised AU recognition experiments on the BP4D database.}
\label{table:bp4d_auc_acc}
\centering
\begin{threeparttable}
\tabcolsep=3pt
\scalebox{0.7}[0.75]{
\begin{tabular}{c|ccc|ccccc|ccccc}\toprule
& \multicolumn{3}{c|}{\textbf{training/development AUC}} & \multicolumn{5}{c|}{\textbf{3-fold AUC}} & \multicolumn{5}{c}{\textbf{3-fold Accuracy}} \\ \cmidrule{2-14}
AU & Ours & ``O-wlc'' & \begin{tabular}[c]{@{}c@{}}\scriptsize OFS-CNN\\\cite{han2018optimizing}\end{tabular} &
Ours & ``O-wlc'' &
\begin{tabular}[c]{@{}c@{}}\scriptsize SRERL\\\cite{li2019semantic}\end{tabular} &
\begin{tabular}[c]{@{}c@{}}\scriptsize DRML\\\cite{zhao2016deep}\end{tabular} &
\begin{tabular}[c]{@{}c@{}}\scriptsize JPML\\\cite{zhao2015joint}\end{tabular} &
Ours & ``O-wlc'' & 
\begin{tabular}[c]{@{}c@{}}\scriptsize ARL\\\cite{shao2019facial}\end{tabular} &
\begin{tabular}[c]{@{}c@{}}\scriptsize JAA-Net\\\cite{shao2018deep}\end{tabular} &
\begin{tabular}[c]{@{}c@{}}\scriptsize EAC-Net\\\cite{li2018eac}\end{tabular} \\ \midrule
1    & \textbf{64.9} & 57.3 &  -   & \textbf{70.4} & 68.4 & 67.6 & 55.7 & 40.7 & 74.1 & 73.9 & 73.9 & \textbf{74.7} & 68.9 \\
2    & \textbf{63.9} & 59.5 &  -   & 67.8 & 65.7 & \textbf{70.0} & 54.5 & 42.1 & 78.3 & 78.0 & 76.7 & \textbf{80.8} & 73.9 \\
4    & \textbf{72.0} & 67.8 &  -   & 74.4 & \textbf{75.3} & 73.4 & 58.8 & 46.2 & 80.1 & \textbf{82.6} & 80.9 & 80.4 & 78.1 \\
6    & \textbf{80.4} & 79.1 &  -   & \textbf{80.2} & 78.5 & 78.4 & 56.6 & 40.0 & \textbf{79.8} & 77.9 & 78.2 & 78.9 & 78.5 \\
7    & \textbf{75.2} & 71.9 &  -   & 75.0 & 74.7 & \textbf{76.1} & 61.0 & 50.0 & \textbf{76.4} & 76.3 & 74.4 & 71.0 & 69.0 \\
10   & \textbf{81.1} & 77.6 &  -   & 76.8 & 78.6 & \textbf{80.0} & 53.6 & 75.2 & 79.5 & \textbf{80.6} & 79.1 & 80.2 & 77.6 \\
12   & \textbf{84.9} & 82.8 &  -   & \textbf{86.3} & 86.0 & 85.9 & 60.8 & 60.5 & \textbf{86.7} & 86.5 & 85.5 & 85.4 & 84.6 \\
14   & \textbf{69.4} & 64.7 &  -   & \textbf{66.7} & 64.9 & 64.4 & 57.0 & 53.6 & \textbf{66.4} & 65.0 & 62.8 & 64.8 & 60.6 \\
15   & \textbf{67.9} & 67.6 &  -   & 70.8 & 70.2 & \textbf{75.1} & 56.2 & 50.1 & 84.1 & 84.0 & \textbf{84.7} & 83.1 & 78.1 \\
17   & 67.7 & 68.2 &  -   & 69.3 & 68.6 & \textbf{71.7} & 50.0 & 42.5 & 70.3 & 69.6 & \textbf{74.1} & 73.5 & 70.6 \\
23   & \textbf{65.4} & 60.4 &  -   & 69.0 & 68.4 & \textbf{71.6} & 53.9 & 51.9 & 82.1 & 81.2 & \textbf{82.9} & 82.3 & 81.0 \\
24   &  -   &  -   &  -   & 68.2 & 67.3 & \textbf{74.6} & 53.9 & 53.2 & 85.6 & 85.7 & \textbf{85.7} & 85.4 & 82.4 \\ \midrule
Avg. & 72.1 & 68.8 & \textbf{72.2} & 72.9 & 72.2 & \textbf{74.1} & 56.0 & 50.5 & \textbf{78.6} & 78.4 & 78.2 & 78.4 & 75.2 \\
\bottomrule
\end{tabular}}
\end{threeparttable}
\end{table*}

%%%%%%%%%%%%%%%%%%%%%%%%%%%%%%%%% DISFA F1 score %%%%%%%%%%%%%%%%%%%%%%%%%%%%%%%%%
%%%
\begin{table*}[!htbp]
\caption{F1 score, AUC and Accuracy of fully supervised AU recognition experiments on the DISFA database.}
\label{table:disfa_f1_auc_acc}
\centering
\begin{threeparttable}
\tabcolsep=2pt
\scalebox{0.7}[0.75]{
\begin{tabular}{c|cccccccccccc|ccccc|ccccc}\toprule
& \multicolumn{12}{c|}{\textbf{3-fold F1}} & \multicolumn{5}{c|}{\textbf{3-fold AUC}} & \multicolumn{5}{c}{\textbf{3-fold Accuracy}} \\ \cmidrule{2-23}
AU & Ours & ``O-wlc'' & 
\begin{tabular}[c]{@{}c@{}}\scriptsize SRERL\\\cite{li2019semantic}\end{tabular} &
\begin{tabular}[c]{@{}c@{}}\scriptsize ARL\\\cite{shao2019facial}\end{tabular} &
\begin{tabular}[c]{@{}c@{}}\scriptsize LP-Net\\\cite{niu2019local}\end{tabular} &
\begin{tabular}[c]{@{}c@{}}\scriptsize U-Net\\\cite{sankaran2020domain}\end{tabular} &
\begin{tabular}[c]{@{}c@{}}\scriptsize DSIN\\\cite{corneanu2018deep}\end{tabular} &
\begin{tabular}[c]{@{}c@{}}\scriptsize JAA-Net\\\cite{shao2018deep}\end{tabular} &
\begin{tabular}[c]{@{}c@{}}\scriptsize ROI\\\cite{li2017action}\end{tabular} &
\begin{tabular}[c]{@{}c@{}}\scriptsize EAC-Net\\\cite{li2018eac}\end{tabular} &
\begin{tabular}[c]{@{}c@{}}\scriptsize DRML\\\cite{zhao2016deep}\end{tabular} &
\begin{tabular}[c]{@{}c@{}}\scriptsize APL\\\cite{zhong2014learning}\end{tabular} &
Ours & ``O-wlc'' &
\begin{tabular}[c]{@{}c@{}}\scriptsize SRERL\\\cite{li2019semantic}\end{tabular} &
\begin{tabular}[c]{@{}c@{}}\scriptsize DRML\\\cite{zhao2016deep}\end{tabular} &
\begin{tabular}[c]{@{}c@{}}\scriptsize APL\\\cite{zhong2014learning}\end{tabular} &
Ours & ``O-wlc'' & 
\begin{tabular}[c]{@{}c@{}}\scriptsize ARL\\\cite{shao2019facial}\end{tabular} &
\begin{tabular}[c]{@{}c@{}}\scriptsize JAA-Net\\\cite{shao2018deep}\end{tabular} &
\begin{tabular}[c]{@{}c@{}}\scriptsize EAC-Net\\\cite{li2018eac}\end{tabular} \\ \midrule
1    & \textbf{51.7} & 46.4 & 45.7 & 43.9 & 29.9 & 44.8 & 46.9 & 43.7 & 41.5 & 41.5 & 17.3 & 11.4 & 76.6 & \textbf{77.7} & 76.2 & 53.3 & 32.7 & \textbf{94.8} & 93.0 & 92.1 & 93.4 & 85.6 \\
2    & 35.1 & 33.6 & \textbf{47.8} & 42.1 & 24.7 & 41.7 & 42.5 & 46.2 & 26.4 & 26.4 & 17.7 & 12.0 & 70.1 & 70.8 & \textbf{80.9} & 53.2 & 27.8 & 92.8 & 91.8 & 92.7 & \textbf{96.1} & 84.9 \\
4    & 67.9 & 66.2 & 59.6 & 63.6 & \textbf{72.7} & 52.9 & 68.8 & 56.0 & 66.4 & 66.4 & 37.4 & 30.1 & \textbf{81.7} & 81.2 & 79.1 & 60.0 & 37.9 & \textbf{89.9} & 89.2 & 88.5 & 86.9 & 79.1 \\
6    & 46.1 & 45.9 & 47.1 & 41.8 & 46.8 & \textbf{57.9} & 32.0 & 41.4 & 50.7 & 50.7 & 29.0 & 12.4 & 69.9 & 68.7 & \textbf{80.4} & 54.9 & 13.6 & 92.0 & \textbf{92.5} & 91.6 & 91.4 & 69.1 \\ 
9    & 56.2 & 49.0 & 45.6 & 40.0 & 49.6 & 50.7 & 51.8 & 44.7 &  8.5 & \textbf{80.5} & 10.7 & 10.1 & 74.5 & 69.7 & \textbf{76.5} & 51.5 & 64.4 & \textbf{96.7} & 96.5 & 95.9 & 95.8 & 88.1 \\
12   & 77.0 & 77.2 & 73.5 & 76.2 & 72.9 & 72.4 & 73.1 & 69.6 & 89.3 & \textbf{89.3} & 37.7 & 65.9 & 88.9 & 88.3 & 87.9 & 54.6 & \textbf{94.2} & 93.6 & 93.8 & \textbf{93.9} & 91.2 & 90.0 \\
25   & 92.5 & 91.9 & 84.3 & \textbf{95.2} & 93.8 & 82.2 & 91.9 & 88.3 & 88.9 & 88.9 & 38.5 & 21.4 & \textbf{94.5} & 93.8 & 90.9 & 45.6 & 50.4 & 95.9 & 95.6 & \textbf{97.3} & 93.4 & 80.5 \\
26   & 45.1 & 47.5 & 43.6 & \textbf{66.8} & 65.0 & 60.9 & 46.6 & 58.4 & 15.6 & 15.6 & 20.1 & 26.9 & 68.7 & 70.6 & \textbf{73.4} & 45.3 & 47.1 & 91.1 & 91.1 & \textbf{94.3} & 93.2 & 64.8 \\ \midrule
Avg. & \textbf{58.9} & 57.2 & 55.9 & 58.7 & 56.9 & 57.9 & 56.7 & 56.0 & 48.5 & 48.5 & 26.7 & 23.8 & 78.1 & 77.6 & \textbf{80.7} & 52.3 & 46.0 & \textbf{93.3} & 92.9 & 93.3 & 92.7 & 80.6 \\
\bottomrule
\end{tabular}}
\end{threeparttable}
\end{table*}
%%%

%%% �Ա�DAU-R��DSIN SRERL)
First, we compare our method to DAU-R, DSIN and SRERL, which also take AU relations into account. We can observe that the proposed method outperforms the three methods on both databases. Specifically, on the BP4D database, our method achieves 9.6\%, 1.6\% and 0.4\% improvement over DAU-R, DSIN and SRERL on F1 score. On the DISFA database, the proposed method achieves 2.2\% and 3.0\% higher F1 score than DSIN and SRERL.
All these methods attempt to make the distribution of the recognized AUs converge with the distribution of the ground truth AUs while minimizing the recognized error between the recognized AUs and the ground truth AUs. DAU-R uses a RBM to model the existing distribution of ground truth AU labels, and forces the network to output predictions that are subject to the learned label distribution. SRERL utilizes AU relationship graph to represent the global relationships between AUs. The AU distributions take a certain form, which may not be identical to the true distribution forms inherent in AU labels due to facial muscle interactions. The proposed method uses adversarial learning to directly force statistical similarity of the distribution of the predicted AU labels and the distribution inherent in ground-truth AU labels, without any assumption of distribution form. Therefore, the proposed method can more completely and faithfully capture inherent AU distributions, resulting in better performance on AU recognition. Moreover, DSIN used local features and global features to train graphic models for learning AU relations. In fact, our method only involves global features, and has achieved better results. It also demonstrates the effectiveness of the proposed method.

%%% �Ա�����ģ�� ARL, LP-Net, U-net
Secondly, the proposed method outperforms other deep models,i.e. ARL, LP-Net, U-Net, JAA-Net, ROI, EAC-Net, OFS-CNN, DRML.
On the BP4D database, our method achieves an average F1 score of 0.610 when examining AUs it shares with OFS-CNN, which is 7.3\% higher than the average F1 score achieved by that model. The proposed method achieves a 15.0\% improvement in average F1 score and 16.9\% improvement in average AUC over DRML. The value of average F1 score of the proposed method is 6.9\%, 7.4\%, 2.7\%, 2.3\% and 2.2\% higher than ROI, EAC-Net, U-Net, LP-Net and ARL, respectively. Our method is also 3.3\% better than JAA-Net on the BP4D database. On the DISFA database, our method still performs better than these methods. In fact, the majority of these methods used multi-label cross entropy to handle multiple AU recognition. Since multi-label cross entropy is the sum of binary cross entropy of each label, it cannot effectively explore label dependencies. The proposed method uses adversarial learning to successfully explore AU distribution from the data. 
% In addition, JAA-Net has tried to enhance the performance of AU detection by attention mechanism and landmark auxiliary. And on the DISFA database, the proposed method achieves a 2.9\% improvement in average F1 score over JAA-Net. The comparable results also show the effectiveness of our method. In total, the superiority of the proposed method to state-of-the-art deep works further proves the crucial role of label distribution for AU recognition.

%%% �Ա�ǳ��ģ��
Thirdly, as expected, the proposed method achieves better performances than the shallow models, JPML. Specifically, the proposed method achieves a 17.4\% improvement in average F1 score and a 22.4\% improvement in average AUC over JPML on the BP4D database. JPML considers positive correlations and negative competitions by introducing constraints in the loss function. The superior performance of the proposed method demonstrates the superiority of deep learning and the potential of the proposed adversarial learning method in capturing high-dimensional distributions from AU data.

%On the DISFA database, only DAU-R used the same experimental conditions as our own. From Table \ref{table:bp4d_auc_results}, we find that the proposed method performs better on most AUs than DAU-R, and the average F1 score of the AUs is 0.593, which is 0.9\% higher than DAU-R. This demonstrates that the adopted adversarial learning technique can more completely and thoroughly capture AU distributions than state-of-the-art generative or discriminative methods, including RBM.
%
%The number of AUs and samples used in GPDE, MCLVM, HRBM, and $l_p$-MTMKL differ from our own. Therefore, the comparisons with these four works are only for reference. In most cases, the proposed method has a higher average F1 score than related works. Specifically, for the AUs we shared with GPDE, the average F1 score of our method is 0.721, which is 11.5\% higher than that of GPDE. For the AUs in common with \cite{eleftheriadis2015multi}, the average F1 score of our method is 0.557, which is 3.1\% higher than that of HRBM and 5.7\% higher than that of $l_p$-MTMKL. This is more evidence of the superiority of the proposed method.

% \subfloat[co-occurrence]{\includegraphics[height=1.5in]{resources/co_exist.jpg}} \hspace{0.4in}
% \subfloat[exclusion]{\includegraphics[height=1.5in]{resources/exclusion.jpg}}

\subsubsection{Evaluation of the Parameter $\alpha$}

% \begin{figure}[!htbp]
% \centering
% \includegraphics[width=3.2in]{resources/bp4d_alpha.png}
% \includegraphics[width=3.2in]{resources/disfa_alpha.png}
% \caption{Impact of parameter $\alpha$ on performance.}
% \label{fig:param_alpha}
% \end{figure}

% \begin{figure}[!tbp]
% \centering
% \subfigure{
% \begin{minipage}{2.4in}
% \centering
% \includegraphics[height=1.6in]{resources/bp4d_alpha.png}
% \end{minipage}
% } \hspace{0.3in}
% \subfigure{
% \begin{minipage}{2.4in}
% \centering
% \includegraphics[height=1.6in]{resources/disfa_alpha.png}
% \end{minipage}
% } 
% \caption{Impact of parameter $\alpha$ on performance.}
% \label{fig:param_alpha}
% \end{figure}

\begin{figure}[!tbp]
\centering
\subfloat[]{\includegraphics[height=1.5in]{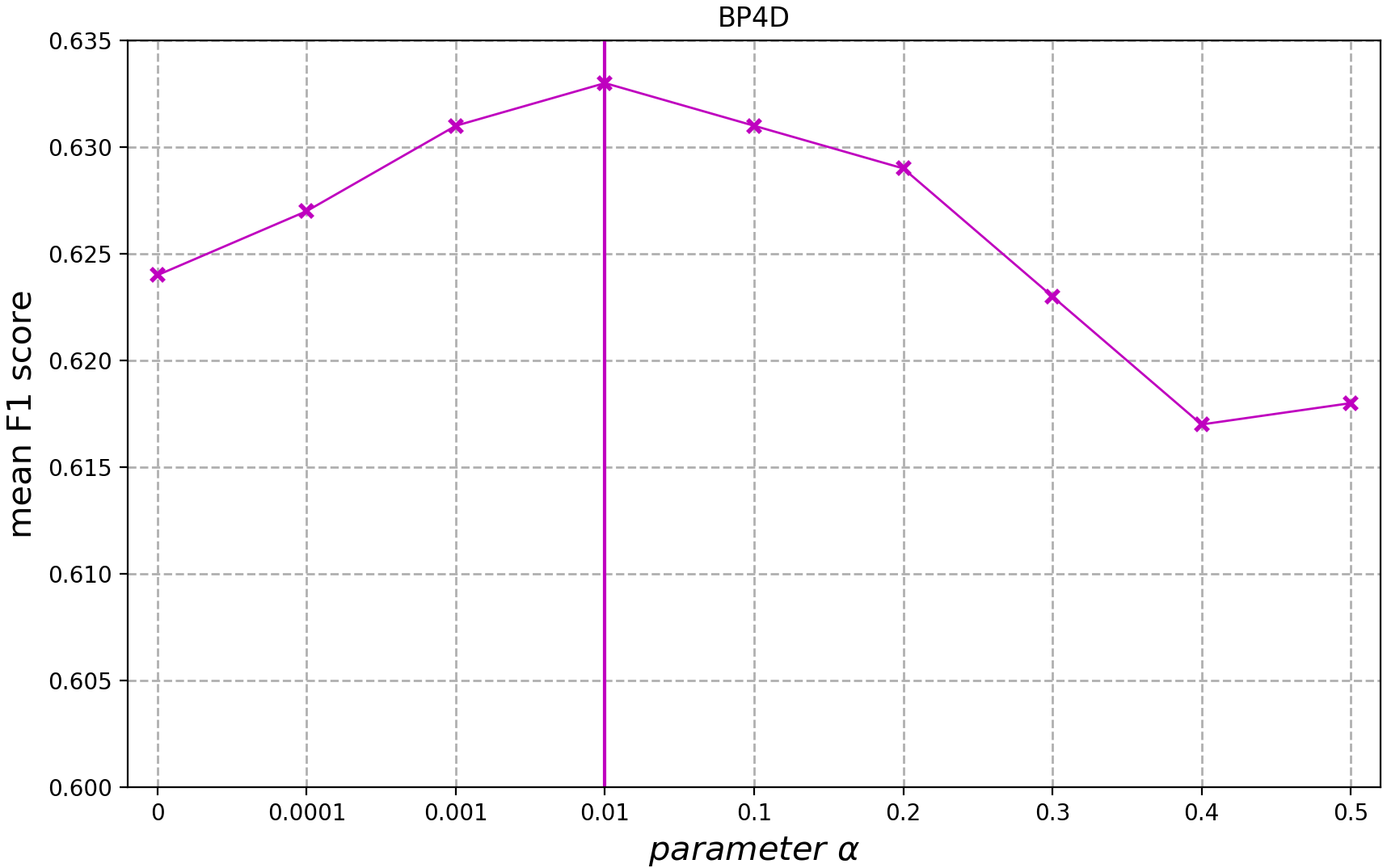}} \hspace{0.4in}
\subfloat[]{\includegraphics[height=1.5in]{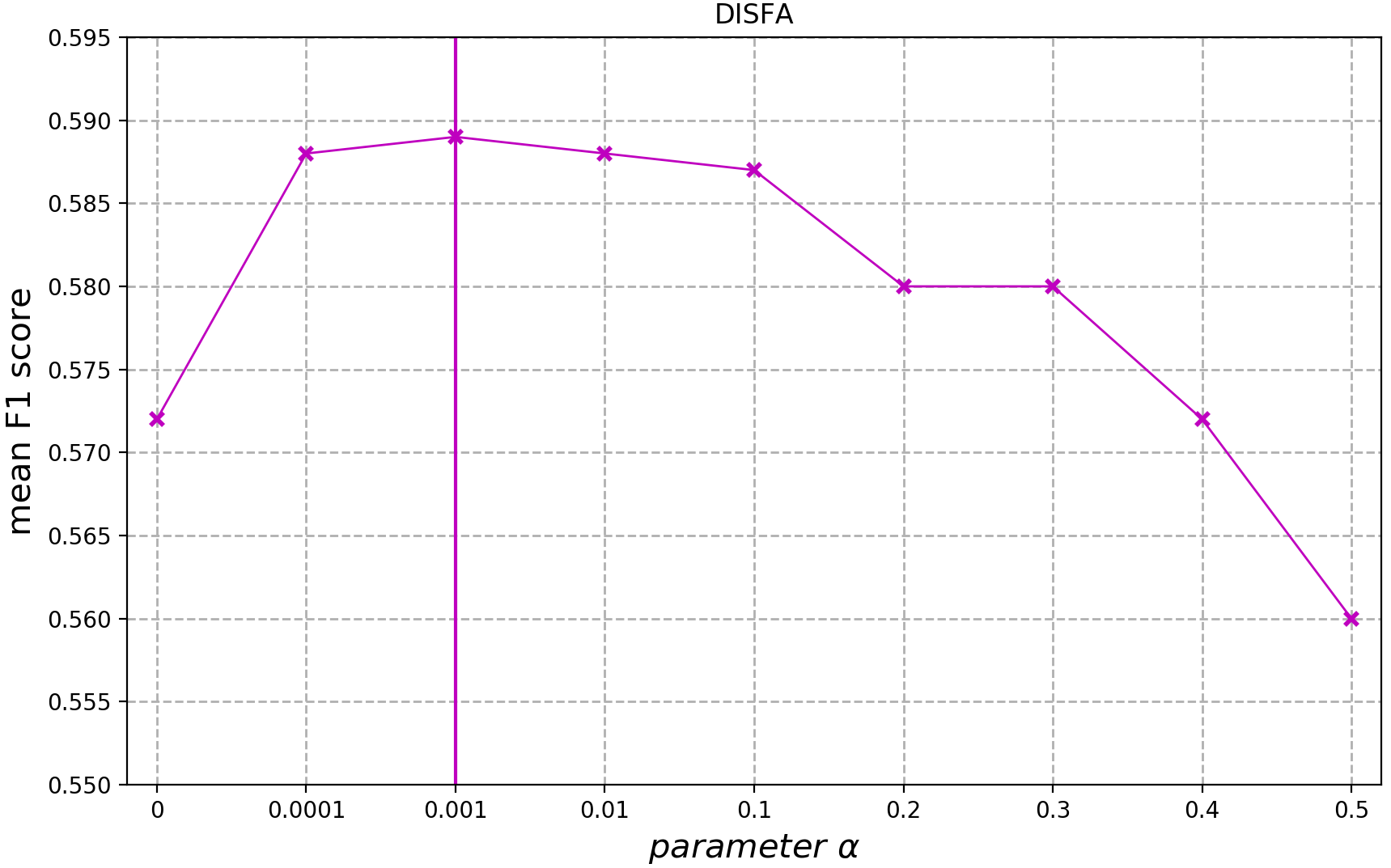}}
\caption{Impact of parameter $\alpha$ on performance.}
\label{fig:param_alpha}
\end{figure}

The parameter $\alpha$ controls the tradeoff between the supervised loss and the adversarial loss. We conduct experiments with different values of $\alpha$ as shown in Figure \ref{fig:param_alpha}. We find an optimal value of $\alpha$ which corresponds to the best tradeoff between supervised and adversarial loss. Specifically, the optimal value of $\alpha$ is 0.01 on the BP4D database and 0.001 on the DISFA database.
The ratio of the adversarial loss and supervised loss is roughly 1:15 on the BP4D database and 1:3 on the DISFA database. This indicates that supervised loss is more important than adversarial loss.

\subsubsection{Evaluation of Learned Feature Representations}

% \begin{figure}[!tbp]
% \centering
% \subfigure{
% \begin{minipage}{2.3in}
% % \centering
% \includegraphics[height=2.0in]{resources/BP4D_raw.pdf}
% \end{minipage}
% } \hspace{0.3in}
% \subfigure{
% \begin{minipage}{2.3in}
% % \centering
% \includegraphics[height=2.0in]{resources/feature_tnse.png}
% \end{minipage}
% } 
% \caption{A t-SNE embedding plotting on the BP4D database. Left: feature space of facial images. Right: feature space of vectors in penultimate layer.}
% \label{fig:tsne}
% \end{figure}

\begin{figure}[!tbp]
\centering
\subfloat[]{\includegraphics[height=1.5in]{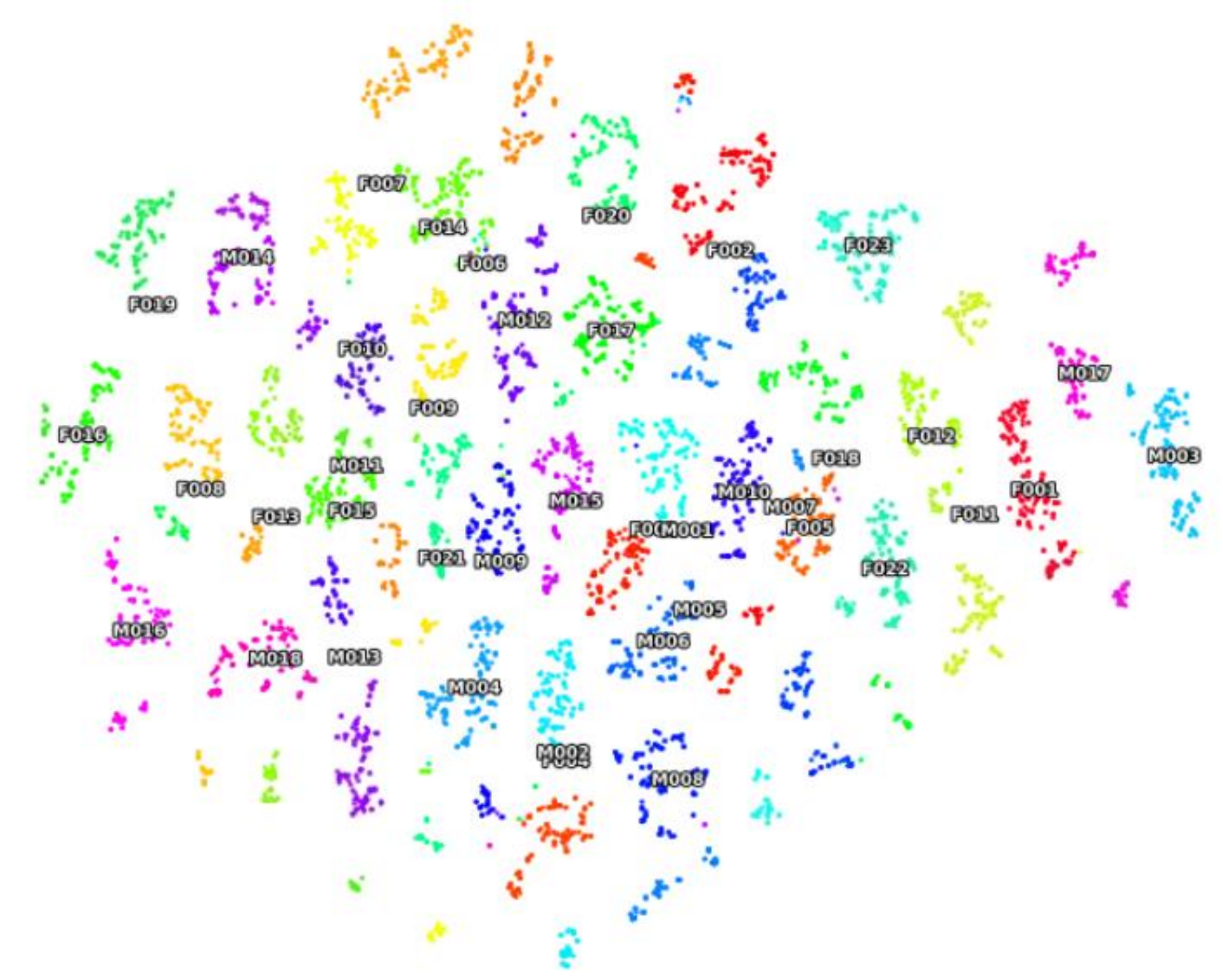}} \hspace{0.4in}
\subfloat[]{\includegraphics[height=1.5in]{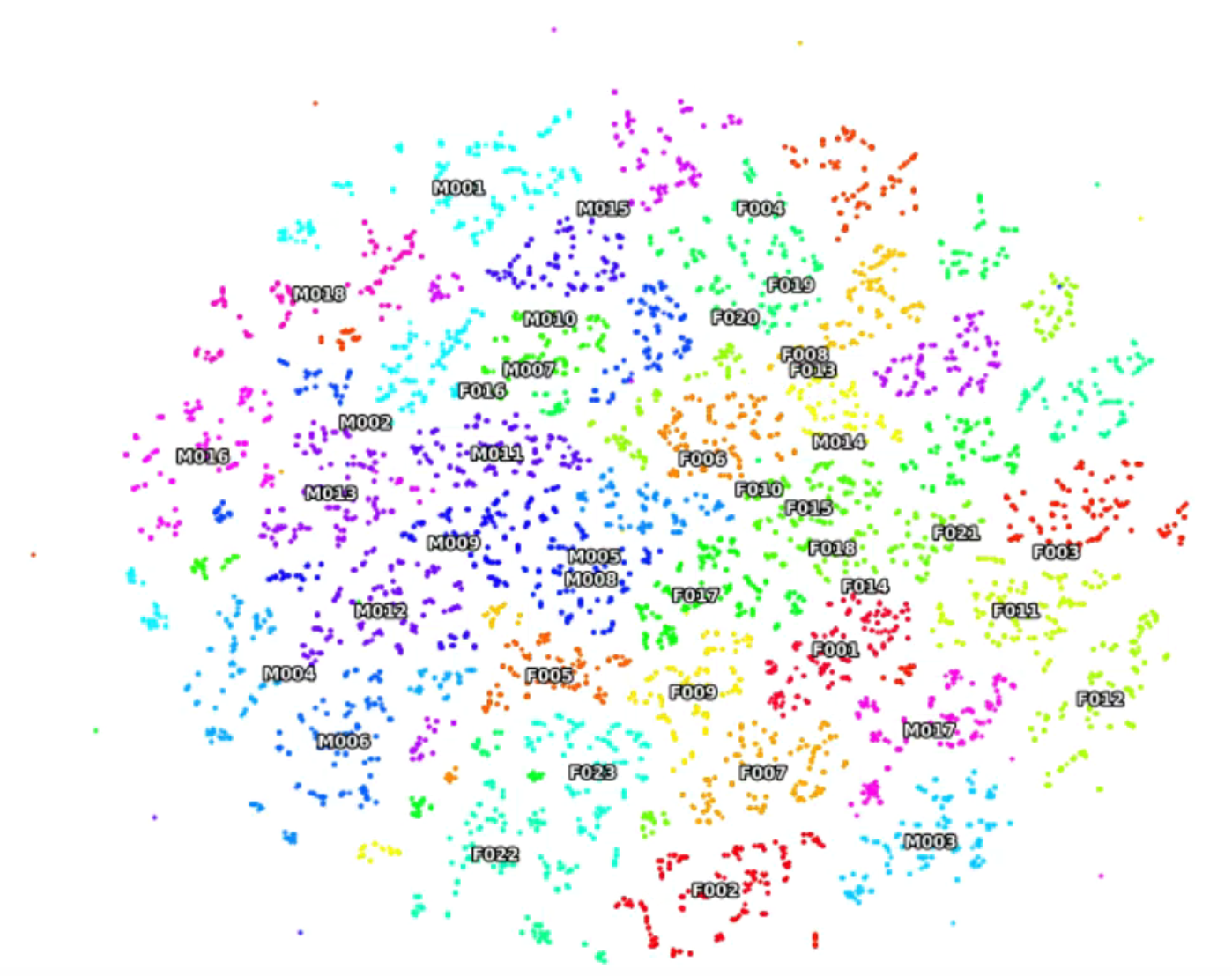}}
\caption{A t-SNE embedding plotting on the BP4D database. (a): feature space of facial images. (b): feature space of vectors in penultimate layer.}
\label{fig:tsne}
\end{figure}

We demonstrate the effectiveness of feature representations by investigating and visualizing the representations in 2D plots with t-SNE \cite{maaten2008visualizing}. Two feature spaces are visualized. The first is the feature space of facial images. The other is the space of the ResNet50 encoding vectors. Figure \ref{fig:tsne} shows the visualization of AU24 on the BP4D database. Samples from the same subject are in the same color and are tagged with the subject ID at the centroid position. From Figure \ref{fig:tsne}, we can see that centroids of different subjects are dispersive in the space of facial images, indicating a high variance between individuals. Compared to the facial image space, samples from different subjects tend to blend together in the learned feature space. This demonstrates that feature representations learned by deep neural networks are not impacted by subject, and are beneficial for the task of AU recognition.

% \begin{figure}[!htbp]
% \centering
% \includegraphics[width=3.0in]{resources/BP4D_raw.pdf}
% \includegraphics[width=3.3in]{resources/feature_tnse.png}
% \caption{A t-SNE embedding plotting on the BP4D database. Up: feature space of facial images. Down: feature space of vectors in penultimate layer.}
% \label{fig:tsne}
% \end{figure}

\section{Conclusion}
Current approaches to facial action unit recognition are primarily based on deep neural networks trained in a supervised manner, and require several labeled facial images for training. Dependencies among AUs are either ignored or captured in a limited fashion. In this paper, we explore adversarial learning for deep semi-supervised facial action unit recognition. We first utilize a deep neural network to learn feature representations and make predictions. Then, a discriminator is introduced to distinguish predicted AUs from ground truth labels. By optimizing the deep AU classifier and the discriminator in an adversarial manner, the deep neural network is able to make predictions which are increasingly close to the actual AU distribution. Experiments on the DISFA and BP4D databases demonstrate that the proposed approach can learn the AU distributions more precisely than state-of-the-art AU recognition methods, outperforming them in both supervised and semi-supervised learning scenarios.

\begin{acks}
This work has been supported by the National Natural Science Foundation of China (Grant No. 917418129, 61727809), and the major project from Anhui Science and Technology Agency (1804a09020038).
\end{acks}

\bibliographystyle{ACM-Reference-Format}
\bibliography{sample-base}

\end{document}